%% file: iclr2024_conference.tex
\documentclass{article} 
\usepackage{subcaption}
\usepackage{hyperref}
\usepackage{url}
\usepackage[usenames,dvipsnames]{xcolor}
\usepackage{amsmath}
\usepackage{wrapfig}
\usepackage{adjustbox}
\usepackage[utf8]{inputenc} 
\usepackage[T1]{fontenc}    
\usepackage{url}            
\usepackage{booktabs}       
\usepackage{amsfonts}       
\usepackage{nicefrac}       
\usepackage{microtype}      
\usepackage{xcolor}         
\usepackage{multirow}
\usepackage{graphicx}
\usepackage{caption}
\usepackage{stfloats}
\usepackage{longtable}
\usepackage[most]{tcolorbox}

\usepackage[preprint]{neurips_2023}

\usepackage{hyperref}
\usepackage{url}

\title{GPT-4V(ision) as a Generalist Evaluator for Vision-Language Tasks}

\author{Xinlu Zhang$^{\diamondsuit*}$,~~Yujie Lu$^{\diamondsuit*}$,~~Weizhi Wang$^{\diamondsuit*}$,~~An Yan$^\spadesuit$,~~Jun Yan$^\clubsuit$,~~Lianke Qin$^\diamondsuit$\\
\textbf{Heng Wang$^\heartsuit$,~~Xifeng Yan$^\diamondsuit$,~~William Yang Wang$^\diamondsuit$, ~~Linda Ruth Petzold$^\diamondsuit$}\\
$^\diamondsuit$University of California, Santa Barbara~~$^\spadesuit$University of California, San Diego  \\
$^\clubsuit$University of Southern California~~$^\heartsuit$Bytedance, US \\
\texttt{~\{xinluzhang, yujielu, weizhiwang, lianke\}@ucsb.edu}\\
\texttt{~anyan@eng.ucsd.edu, yanjun@usc.edu}\\
}

\begin{document}

\definecolor{ourblue}{HTML}{7BB0DF}
\maketitle
\def\thefootnote{*}\footnotetext{Equal Contribution}\def\thefootnote{\arabic{footnote}}

\begin{abstract}
Automatically evaluating vision-language tasks is challenging, especially when it comes to reflecting human judgments due to limitations in accounting for fine-grained details. Although GPT-4V has shown promising results in various multi-modal tasks, leveraging GPT-4V as a generalist evaluator for these tasks has not yet been systematically explored. We comprehensively validate GPT-4V's capabilities for evaluation purposes, addressing tasks ranging from foundational image-to-text and text-to-image synthesis to high-level image-to-image translations and multi-images to text alignment. We employ two evaluation methods, single-answer grading and pairwise comparison, using GPT-4V. Notably, GPT-4V shows promising agreement with humans across various tasks and evaluation methods, demonstrating immense potential for multi-modal LLMs as evaluators. Despite limitations like restricted visual clarity grading and real-world complex reasoning, its ability to provide human-aligned scores enriched with detailed explanations is promising for universal automatic evaluator.\footnote{Work in Progress}

\end{abstract}

\section{Introduction}
\input{subsection/intro}

\section{GPT-4V-as-a-Generalist-Evaluator}
\label{method}

\input{subsection/method}

\section{Evaluation}
\label{sec:evaluation}
To test if GPT-4V can be used as a reliable evaluator, we conduct experiments on four different tasks and aim to answer four main questions:
\begin{enumerate}
\item Can GPT-4V provide a reasonable evaluator with explanation for different multi-modal tasks?
\item Does GPT-4V make consistent predictions across different evaluation methods described in Section \ref{method}?
\item Does GPT-4V match what humans think in single-answer and pairwise evaluation, respectively? 
\end{enumerate}

\subsection{Image-to-Text Captioning}

\input{subsection/exp_i2t}

\subsection{Text-to-Image Generation}
\input{subsection/exp_t2i}

\subsection{Text-guided Image Editing }
\input{subsection/exp_editing}

\subsection{Multiple Images to Text Alignment}
\input{subsection/multi-image}
\section{Related Work}
\input{subsection/relatedwork}

\section{Discussion}
\input{subsection/limitation}

\section{Conclusion}
In this paper, we propose GPT-4V as an evaluator for multimodal tasks and systematically examine its efficacy on four different tasks: image-to-text captioning, text-to-image generation, text-guided image editing, and multiple-image to text alignment. Our results reveal that GPT-4V achieves a promising agreement rate with human evaluators across different tasks, thereby laying a robust foundation for a MLLM-based evaluation framework.

\bibliography{iclr2024_conference}
\bibliographystyle{iclr2024_conference}
\newpage
\appendix
\input{subsection/app}

\end{document}

%% file: subsection/intro.tex
In recent years, significant progress has been made in vision-language research, including advancements in image-to-text generation~\citep{blip2}, text-to-image generation~ \citep{rombach2021highresolution,ramesh2022hierarchical}, and image-to-image translation~ \citep{brooks2023instructpix2pix,zhang2023magicbrush}. Concurrently, there is growing recognition of the importance of automatic metrics in vision-language tasks, as evidenced by recent work~ \citep{hessel-etal-2021-clipscore, hu2023tifa,lu2023llmscore, xu2023imagereward, wu2023human, kirstain2023pickapic}. However, developing automatic metrics that align with human preferences and provide explanatory insights can be challenging. Most existing text-image alignment metrics specialize in providing similarity scores between image and text features, leading to three key limitations: 1) 
\textbf{Reference-free Constraint}: existing specialized scoring models mostly require reference (e.g., CIDEr~\citep{vedantam2015cider}), which is not always applicable in a broad range of tasks (e.g., text-to-image synthesis); 2) \textbf{Lack of Free-form Evaluation}: they do not adhere to specialized evaluation guidelines and fail to offer quantitative scores that satisfy multiple criteria or detailed qualitative explanations, which makes it challenging to align its assessments with human evaluators or even keep consistent with itself.
3) \textbf{Single-Pair Constraint}: they can only evaluate one image-text pair at a time, making it less suitable for more complex tasks that involve multiple image-text pairs, such as text-guided image editing. 
A question comes up: \textit{Is there a promising universal automatic evaluator?}

Recent advancements in Large Language Models (LLMs) have demonstrated strong instruction-following ability \citep{chatgpt, gpt4, insgpt}, being helpful in real-world tasks \citep{llmjudge,zhang2023alpacareinstructiontuned, chen2023alpagasus}. LLM-as-a-judge \citep{llmjudge} adopts LLM as a judge to evaluate outputs generated by different chatbot models. GPT-4 obtains a strong agreement with human preferences in the text evaluations. Though LLM shows promising results as a auto evaluator to substitute human annotator, the research in the multi-modal large language models (MLLMs) for evaluating diverse text-image tasks still under exploration. 

GPT-4V~\citep{gpt4vcontribution} shows strong performance on various tasks~\citep{yang2023dawn}, being able to provide detailed explanations based on customized instructions on different multi-modal tasks by receiving mutltimodal inputs. However, its capability to serve as evaluator for vision-language tasks remains under-explored. In this work, inspired by \cite{llmjudge}, we systematically study the \textit{GPT-4V-as-a-Generalist-Evaluator}, to explore how well GPT-4V aligns with human preference as a evaluator in various vision-language tasks.

To summarize, our main contributions are below:
\begin{itemize}
    \item We systematically validate the capabilities of GPT-4V as the \textit{evaluator} in a broad range of tasks, including image-to-text task (e.g., image captioning), text-to-image task (e.g., text-to-image generation), image-to-image task (e.g., text-guided image editing).
    \item We demonstrate that GPT-4V is generally a good \textit{reference-free} human-aligned evaluator when prompted carefully, which also provides reasonable and fine-grained explanation.
    \item We identify some limitation of GPT-4V when used as an evaluator, such as perceptual evaluation (e.g., vision clarity) and real-world complex cases. 
\end{itemize}

\begin{figure*}[t] 
\centering 
\includegraphics[width=0.9\textwidth]{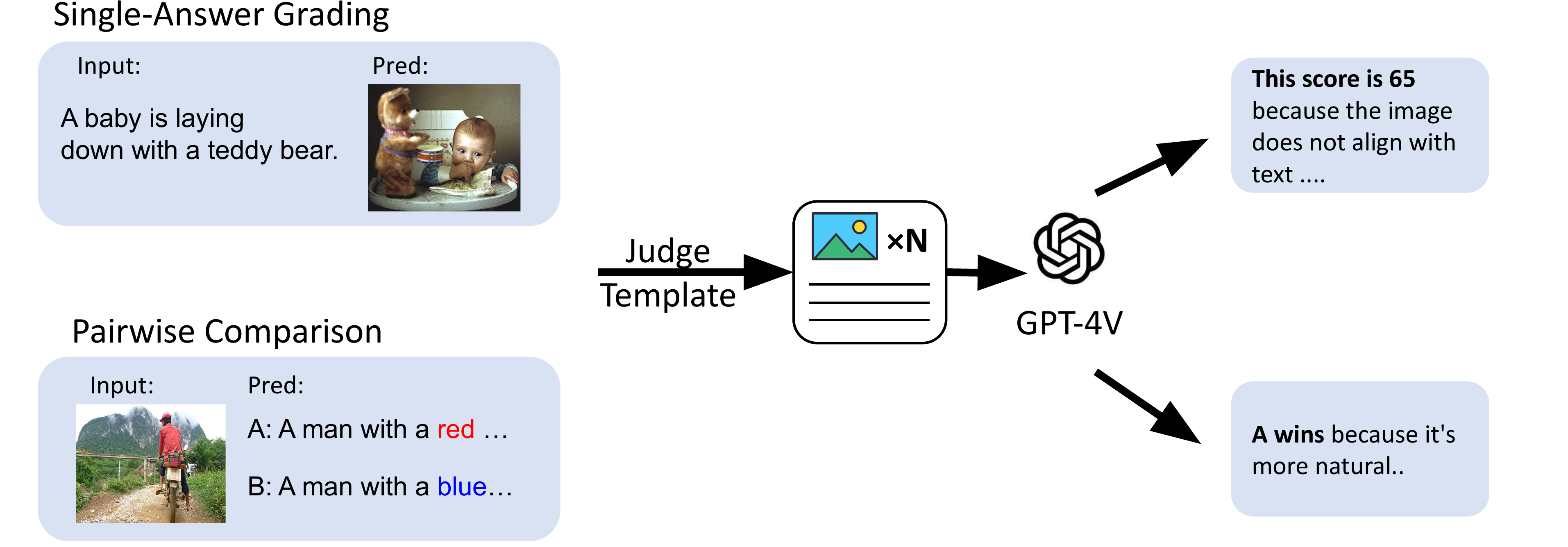}
\vspace{-5pt}
\caption{  \textbf{Overview of GPT-4V Eval.} Our method role GPT-4V as a evaluator to assess the quality of multimodal input-output pairs of different tasks. We adopt two evaluation methods (a) single-answer grading and (b) pairwise comparison, and prompt GPT-4V to evaluate the task.
}
\label{fig:demonstration}
\end{figure*}

%% file: subsection/method.tex
Large Language Model (LLM) has shown promising performance in substituting human annotators for text evaluation tasks~\citep{llmjudge}. Similarly, GPT-4V has demonstrated strong results across a variety of image-based tasks~\citep{yang2023dawn}. However, the role of GPT-4V as a multimodal evaluator remains unexplored. In this section, we demonstrate  how we leverage GPT-4V as a evaluator for evaluating various multimodal tasks.

Inspired by \cite{llmjudge}, we employ two distinct evaluation methods: (1) single-answer grading and (2) pairwise comparison. These methods leverage GPT-4V to assess the quality of outputs across various multimodal tasks, taking into consideration the specific inputs associated with each task. The nature of these input-output pairs varies depending on the type of task. For example, in an image captioning task, the input is an image and the output is text, whereas in synthetic image generation, the roles are reversed\footnote{We conduct different evaluation methods based on the nature of the tasks. Some tasks are evaluated using both single-answer grading and pairwise comparison, while others are evaluated using only one method. We detail each evaluation setting in the corresponding subsection in Section~\ref{sec:evaluation}.}.

\begin{itemize}
  \item{\textbf{Single-answer grading:}  GPT-4V is instructed to generate a score to evaluate the quality and alignment of the input-output pair. The scoring scale may vary depending on the task.}
  \item{\textbf{Pairwise comparison:} GPT-4V evaluates a single task input along with a pair of answers. It makes a evaluation to determine which of the two candidate answers is better, or whether the two answers are of equal quality. }
\end{itemize}

Single-answer grading offers scalability as the number of candidate answers increases. However, it lacks capability to  directly compare different answers for a specific task input. On the other hand, pairwise comparison allows for the direct comparison, while its computational complexity increases quadratically as the number of candidate answers grows. 

\begin{figure*}[t] 
\centering 
\includegraphics[width=0.9\textwidth]{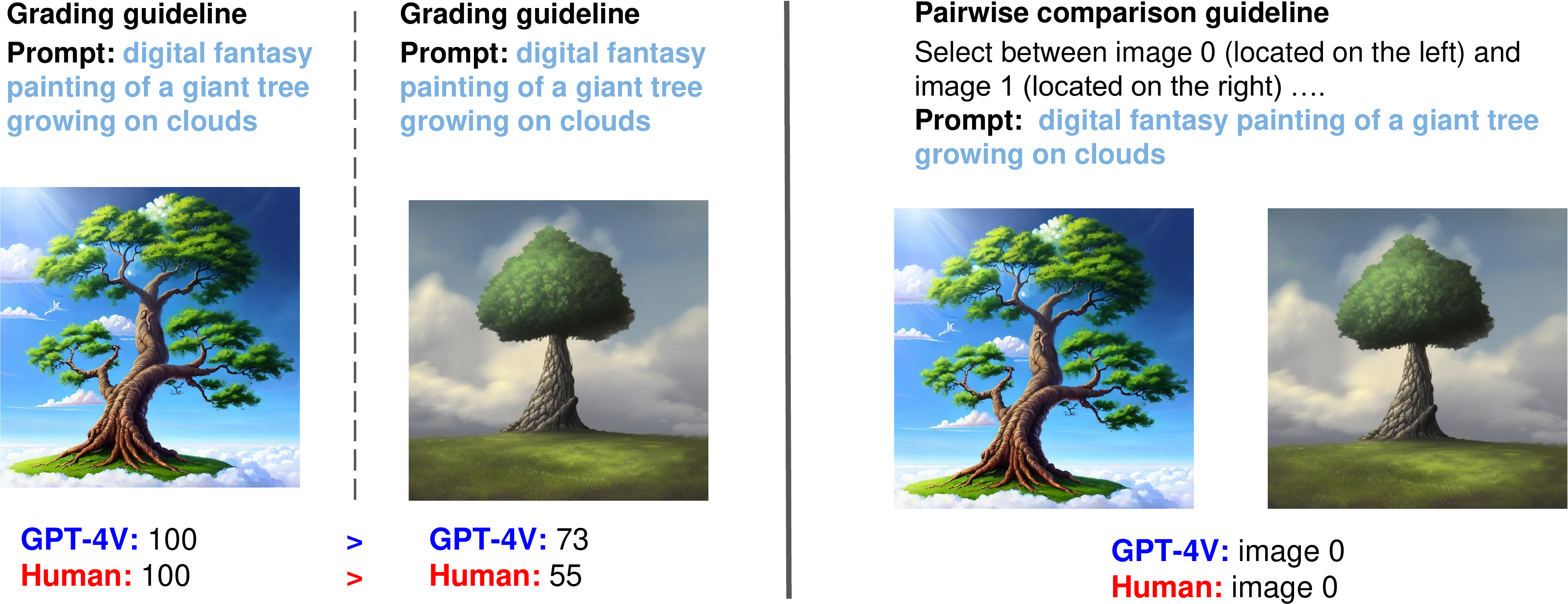} 
\vspace{5pt}
\caption{\textbf{Evaluation methods for single-answer grading (left) and pairwise comparison (right) on text-to-image task.} \textbf{\color{ourblue} lightblue} is the input prompt for image generation, \textbf{\color{blue} blue} and \textbf{\color{red} red} are decisions made by different evaluators.
Left: Evaluators independently evaluate each candidate answer based on the task input. The candidate answer with the higher score is deemed the evaluator's final decision.
Right: Evaluators compare a pair of candidate answers and directly make a final decision. 
}
\label{fig:method_exp}
\end{figure*}

To evaluate the effectiveness of GPT-4V as a evaluator, we conduct a comparative evaluation with human evaluators and GPT-4V. Both human evaluators and GPT-4V assess input-output pairs from various tasks to assess output quality. We categorize the results into ``Answer 1,'' ``Answer 2,'' and ``Tie,'' and calculate alignment scores to measure agreement among different evaluation approaches. We illustrate a single-answer grading and pairwise comparison example in Figure \ref{fig:method_exp}.


\textbf{Single-answer grading.} We ask the evaluator to provide a score within a predefined scale range. The score is based on the multimodal input-output pairs for different tasks. To obtain a final answer that represents the evaluator's preference between two candidate outputs, we consider the one with the higher score as the preferred answer. If the scores are equal, the result is declared a ``Tie''.

\textbf{Pairwise comparison.} 
To ensure a fair assessment, we implement a dual-sided comparing system by using GPT-4V to mitigate the potential position bias \citep{llmjudge}. Each output comparison is evaluated twice, where the answer 1 and the answer 2 are alternated, with the same input task.
To evaluate the final answer in dual-sided pair evaluation we follow:
\begin{itemize}
    \item Answer 1: GPT-4V chooses Answer 1 twice or chooses once Answer 1 and once Tie.
    \item Tie: GPT-4V obtains twice Tie or chooses once Answer 1 and once Answer 2.
    \item Answer 2: GPT-4V chooses Answer 2 twice or chooses once Answer 2 and once Tie.
\end{itemize}

%% file: subsection/exp_i2t.tex
\begin{table}[t]
\centering
\caption{
\textbf{Comparison of grading between human and single-answer grading evaluation for the image-to-text task using both Pearson and Spearman correlations.} Images are scored out of a total of 100. $^{*}$ denotes significant correlations at $p < 0.05$.
}
\begin{tabular}{l llll}
\toprule
 & \multicolumn{2}{c}{Ground-Truth} &  \multicolumn{2}{c}{Hard-Negative } \\
 & Pearson & Spearman & Pearson & Spearman \\
\midrule
CLIPScore  & 0.164 & 0.072 & 0.098 & 0.036  \\
GPT-4V  & 0.401$^*$ & 0.499$^*$ &  0.687$^*$  & 0.612$^*$  \\
\bottomrule
\end{tabular}
 \label{tab:single_score_i2t_results}
\end{table}

\begin{table}[t]
\centering
\caption{\textbf{Agreement [\%] between evaluation methods across GPT-4V and human validates the image-to-text task.} ``GPT-4V-Pair'' and ``GPT-4V-Single'' represent GPT-4V evaluations using pairwise comparison and single-answer grading, respectively. The baseline agreement between two random evaluators is 33\%.}
\begin{tabular}[t]{l c c }
\toprule
Evaluator & GPT-4V-Single & Human\\
\midrule
GPT-4V-Pair & 91.0 & 95.0 \\
GPT-4V-Single& - & 86.0  \\
\bottomrule
\end{tabular}
\vspace{-10pt}
\label{tab:i2t_agreement}
\end{table}

\paragraph{Data Construction.}
We first sample two subsets from MSCOCO Train 2017~\citep{mscoco} and Conceptual Captions 12M~\citep{cc12m} image-text paired datasets for evaluating GPT-4V on the image-to-text generation tasks. For each dataset, we sampled 50 image-text pairs via clustering-based sampling to ensure the diversity in constructed dataset and in total the constructed dataset size is 100. 

\paragraph{Evaluation Setup.} We propose to evaluate whether GPT-4V can be consistent, accurate, and unbiased in providing a score towards the generated caption quality in both single scoring and pairwise comparison settings. Due to the lack of negative captions used for paired comparison with ground truth high-quality caption, we follow ALIGN~\citep{align} to manually construct hard negative captions via adding the following noises: 1) category change, i.e., a Toyota car to a Volkswagen car; 2) color change, i.e., white to gray; 3) location change, i.e., changing the country name; 4) number change, i.e., 20 to 50; 5) position change, i.e., behind to front; 6) subtle dropping, i.e., deleting the adverbial to the object. These introduced noises still keep the noisy caption a fluent sentence and differentiating these noisy captions from original ground-truth ones requires the evaluator to figure out the fine-grained dis-alignment between noisy words and the image details, making such negative captions \textbf{hard} to grade.

We ask expert-level human labelers to give consistent scoring towards both the ground-truth and  hard-negative captions in the same scale of 100. Then we firstly compute the correlation and alignment between GPT-4V's scoring and human expert scoring in terms of image-text alignment. We report Pearson and Spearman correlation scores for both ground-truth set and hard-negative set. We also compare GPT-4V-as-an-evaluator with the strong reference-free baseline CLIPScore~\citep{hessel2021clipscore} on the single-answer grading setting. The CLIPScore is calculated following \citet{hessel2021clipscore} using ViT-L/14~\citep{vit} pre-trained checkpoint and it is scaled to $[0,100]$. Secondly, in terms of the pairwise comparison,  we report the agreement score between GPT-4V's evaluation and human evaluation in both single-answer scoring and pairwise comparison settings. In addition, we also compute the agreement between GPT-4V single-answer scoring and pairwise comparison to demonstrate its consistency in different evaluation variants.

\paragraph{GPT-4V is robust and effective in evaluating image-to-text captioning tasks.} The results on the agreement between human and GPT-4V scoring are presented in Table~\ref{tab:single_score_i2t_results}. GPT-4V's scoring is significantly correlated with human rating in both ground-truth subset evaluation and hard-negative subset evaluation. GPT-4V also significantly outperforms CLIPScore in human grading alignment with much higher correlation scores. We analyze the standard deviation to take insights on the score distribution given by GPT-4V and humans.
We find that the standard deviation of GPT-4V grading is much larger than that of human grading. We attribute such fluctuation around the average score to the nucleus decoding used by GPT-4V. When doing sampling based decoding, a group of neighboring scores like 80/82/85 are all within the sampling candidates and thus the final output scores contains much more randomness than that of humans.

\paragraph{GPT-4V is as consistent as human on different evaluation methods on image-to-text captioning tasks.} The results on the consistency analysis across different evaluation methods are presented in Table~\ref{tab:i2t_agreement}. GPT-4V-Single's label is computed based on the larger score of the ground-truth caption and hard-negative caption towards the same image. Then we compute the agreement between GPT-4V-Pair and GPT-4V-Single to present the consistency across different evaluation methods. We can conclude that GPT-4V is a robust and consistent evaluator on image captioning tasks, which achieves an agreement score of 91\% between single-pair-scoring and pairwise-comparison. Additionally, GPT-4V is also consistent with human scoring using both evaluation methods. The human agreement score of GPT-4V-Single is much lower than that of GPT-4V-Pair, which aligns with human intuition that the direct comparison is easier to complete than keeping a consistent standard on single-answer grading.

\begin{figure*}[t!] 
\centering 
\includegraphics[width=0.9\textwidth]{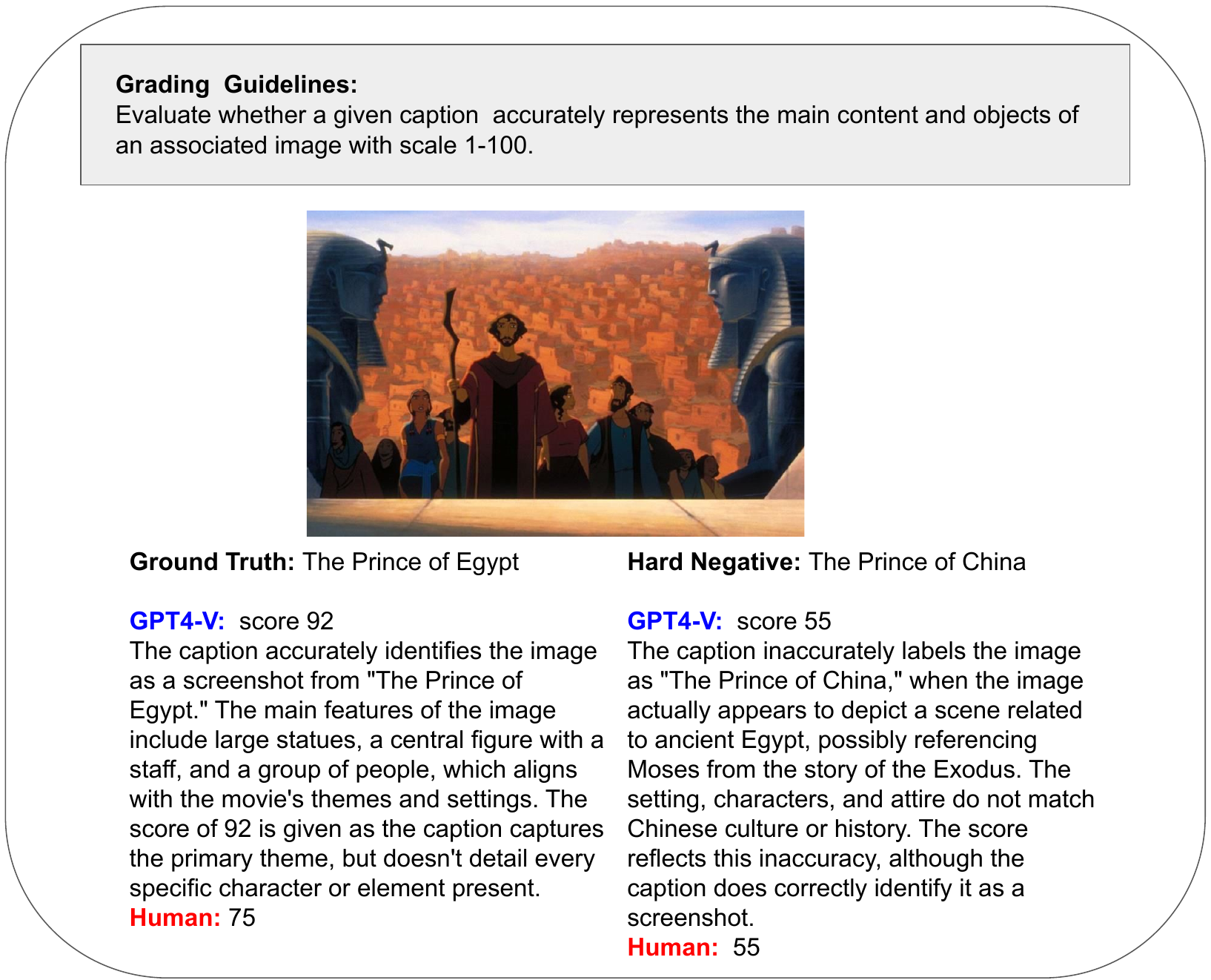} 
\caption{\textbf{Examples of image-to-text evaluation task through single-answer grading on human-preferred ground truth and hard negative outputs, respectively. }GPT-4V successfully identifies the text caption quality and provides comprehensive explanation for its scores.
}
\label{fig:i2t_exp}
\end{figure*}

%% file: subsection/exp_t2i.tex
\begin{table}[t!]
\centering
\caption{
\textbf{Comparison of grading between human and single-answer grading evaluation for the text-to-image task using both Pearson and Spearman correlations.} Images are scored out of a total of 100, divided among three categories: \textit{Relevance} (40 points), \textit{Visual Clarity} (30 points), and \textit{Object Accuracy} (30 points).
 $^{*}$ denotes significant correlations at $p < 0.05$.
}
\begin{tabular}{l l  cccc}
\toprule  
&&Relevance & Visual Clarity & Object Accuracy & Total\\
 \midrule
\multirow{2}{*}{Pearson} & Ground-Truth &  0.530$^{*}$ & 0.247 & 0.451 & 0.502  \\
& Hard-Negative&0.653$^{*}$& 0.318$^{*}$ & 0.587$^{*}$ &0.601$^{*}$   \\
\multirow{2}{*}{Spearman} & Ground-Truth&  0.581$^{*}$  & 0.102 & 0.469$^{*}$  & 0.496$^{*}$  \\
& Hard-Negative& 0.577$^{*}$  & 0.194 & 0.584$^{*}$ & 0.535$^{*}$   \\
\bottomrule
\end{tabular}
 \label{tab:single_score_t2i_results}
\end{table}

\begin{table}[t!]
\centering
\caption{\textbf{Agreement [\%] between evaluation methods across GPT-4V and human validates text-to-image task.} "GPT-4V-Pair" and "GPT-4V-Single" represent GPT-4V evaluations using pairwise comparison and single-answer grading, respectively. The baseline agreement between two random evaluators is 33\%.}
\begin{tabular}[t]{l c c }
\toprule
Evaluator & GPT-4V-Single & Human\\
\midrule
GPT-4V-Pair & 41.43 & 52.85 \\
GPT-4V-Single& -& 51.43  \\
 \bottomrule
\end{tabular}
 \label{tab:t2i_results}
\end{table}

Text-to-image generation involves feeding a text prompt into a model to produce an image that aligns with the specific requirements outlined in the prompt. We introduce GPT-4V as an evaluator to assess pairs of syntactically generated images, determining which image is of higher quality and better aligns with the text prompt. 

\paragraph{Data Construction.} 
We selected 20 samples from ImagenHub ~\citep{ku2023imagenhub} and 50 from Pick-a-Pic \citep{kirstain2023pickapic} for evaluation. ImagenHub offers diverse compositional prompts, while Pick-a-Pic uses real-user prompts that often have multiple requirements, creating a realistic challenge for GPT-4V image synthesis. Each sample contains a text prompt and a pair of images for evaluation. We first manually annotate two images in each sample, one to be ground-truth and the other to be hard negative based on the alignment of the prompt text. It is worth noting that the hard negative samples in text-to-image generation tasks are much \textbf{difficult} to distinguish than that of captioning tasks. It can even take several minutes for a human expert to identify the better one from the good image and the hard-negative one.

\paragraph{Evaluation Setup.}
To ensure a comprehensive and unbiased evaluation in the single-answer grading setting, we propose to assess the generated image from three key dimensions: Relevance, Visual Clarity, and Objective Accuracy, totaling 100 points\footnote{We defined the detailed score criteria into appendix.}. Specifically, 
\begin{itemize}
\item Relevance (40): Alignment of the image and the prompt, including both explicit and implicit details.
\item Visual Clarity (30): Technical quality such as sharpness and absence of artifacts.
\item Objective Accuracy (30): Accuracy of  the objects or nouns described in the prompt, without extraneous elements.
\end{itemize}
Both human and GPT-4V grade each text-image pair based on the three-dimension criteria.
For pair-comparison setting, the GPT-4V is asked to compare the pair of generated images with chain-of-though prompting to improve reasoning capability.

\begin{figure*}[ht!] 
\centering 
\includegraphics[width=0.9\textwidth]{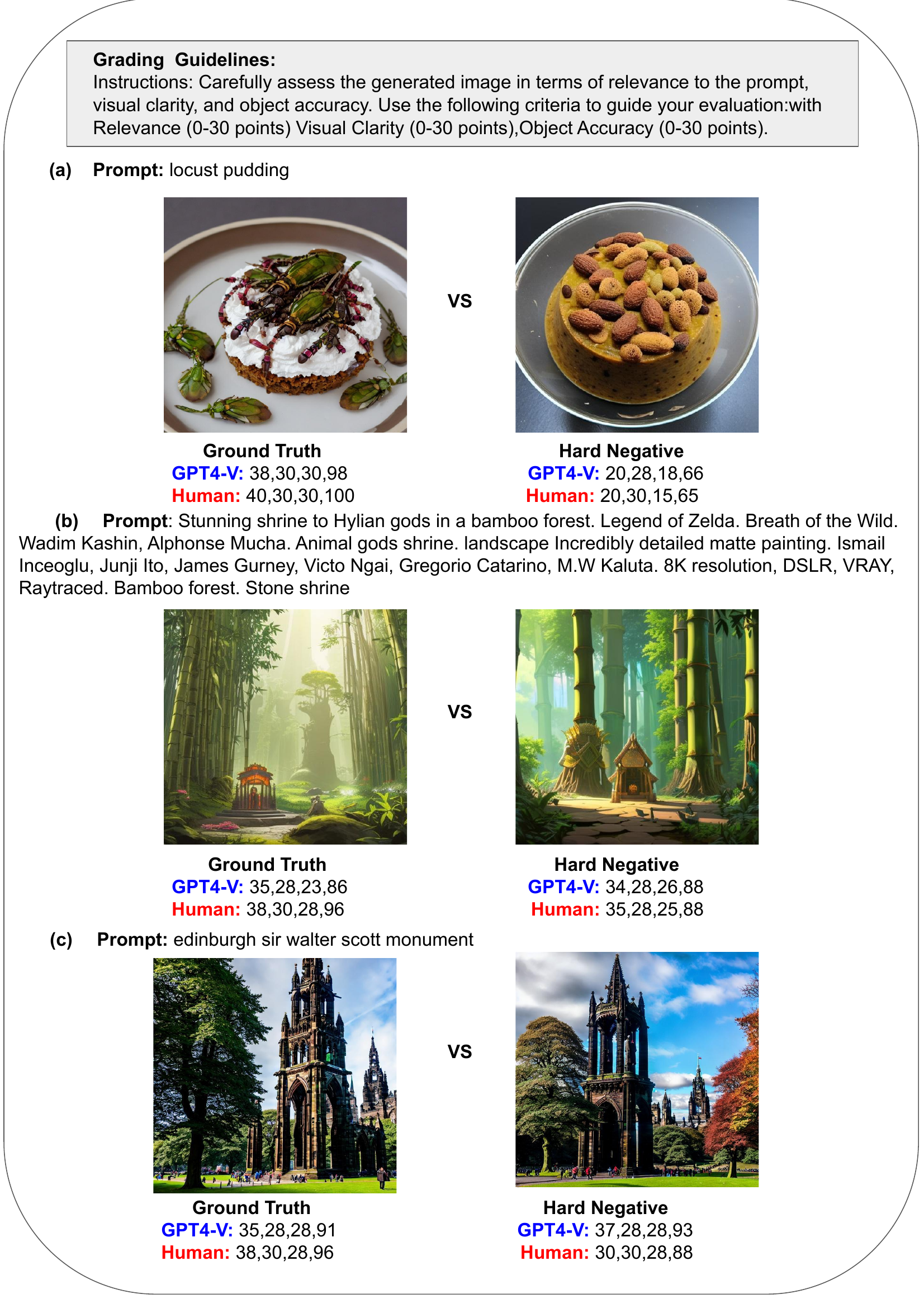} 
\caption{\textbf{Three Examples of  text-to-image evaluation task through single-answer grading on human-preferred ground truth and hard negative outputs, respectively.} (a) GPT-4V successfully identity the image qualities, when the gap between text-image pair is relative large. However, GPT-4V struggles to accurately identify instances where (b) the prompt has multiple stringent requirements, and (c) the evaluation demands a strong foundation of background knowledge. Noting: these challenges are also difficult for human evaluators, requiring meticulous scrutiny and reference to external knowledge for accurate scoring.
}
\vspace{-20pt}
\label{fig:t2i_all}
\end{figure*}

\paragraph{GPT-4V exhibits potentials to score text-to-image tasks.}
We calculate the Person and Spearman correlations and average score of GPT-4V and human across the ground-truth images and hard negative images on three sub grading categories and the total score, the results are shown in the upper part of Table \ref{tab:single_score_t2i_results}.

GPT-4V exhibits a general correlation with human evaluations in terms of the total score for both ground-truth and hard negative images, underscoring the viability of employing GPT-4V for text-to-image evaluation tasks. Delving into specific scoring categories, GPT-4V manifests a moderate correlation with human ratings in relevance and object accuracy for both ground-truth and hard negative images. However, GPT-4V struggles in detecting visual errors in images, tending to award higher scores for visual clarity compared to human evaluators. In addition, we observed that GPT-4V tends to have more statistically significant correlation with humans in the hard negative images. This suggests that while GPT-4V may be less discerning in spotting subtle imperfections, it aligns more closely with human perspectives when grading images with more pronounced flaws or deviations.

\textbf{GPT-4V shows promising agreement with human on different evaluation methods on text-to-image tasks.} To study the agreement between evaluation methods employed by GPT-4V and human evaluators, we compare both single-answer and pairwise evaluations with human. Specifically, we compute the number of agreement between two evaluators and divide the number by the total number of tasks. The results are presented in Table \ref{tab:t2i_results}.

GPT-4V with both single answer grading and pairwise comparison show  promising agreements with human, achieving more than 50\% agreement rate, given the random guess agreement is 33\% of two random evaluators. However, the agreement between single-answer grading and pairwise comparison conducted by GPT-4-V falls short. This discrepancy may arise because the instances we selected are challenging; some are even difficult for humans to assess, requiring nuanced judgment and extensive background knowledge. Different prompts and a larger sample size could affect the agreement rates. We plan to explore this further in future work.

\paragraph{GPT-4V shows capability in evaluating image quality in text-to-image tasks.}

We further study how GPT-4V aligns successfully with human evaluations and identify cases where GPT-4V struggles. We conduct case studies for single-answer grading, shown in Figure \ref{fig:t2i_all}.

GPT-4V aligns with human grades when the image has a relatively large gap compared to the prompt text, as shown in Figure \ref{fig:t2i_all} (a). However, it fails to align with human scores when the prompt text is noisy and contains multiple detailed requirements, as shown in Figure \ref{fig:t2i_all} (b), or requires intensive knowledge, as shown in Figure \ref{fig:t2i_all} (c).

%% file: subsection/exp_editing.tex
\begin{figure*}[t!] 
\centering 
\includegraphics[width=0.9\textwidth]{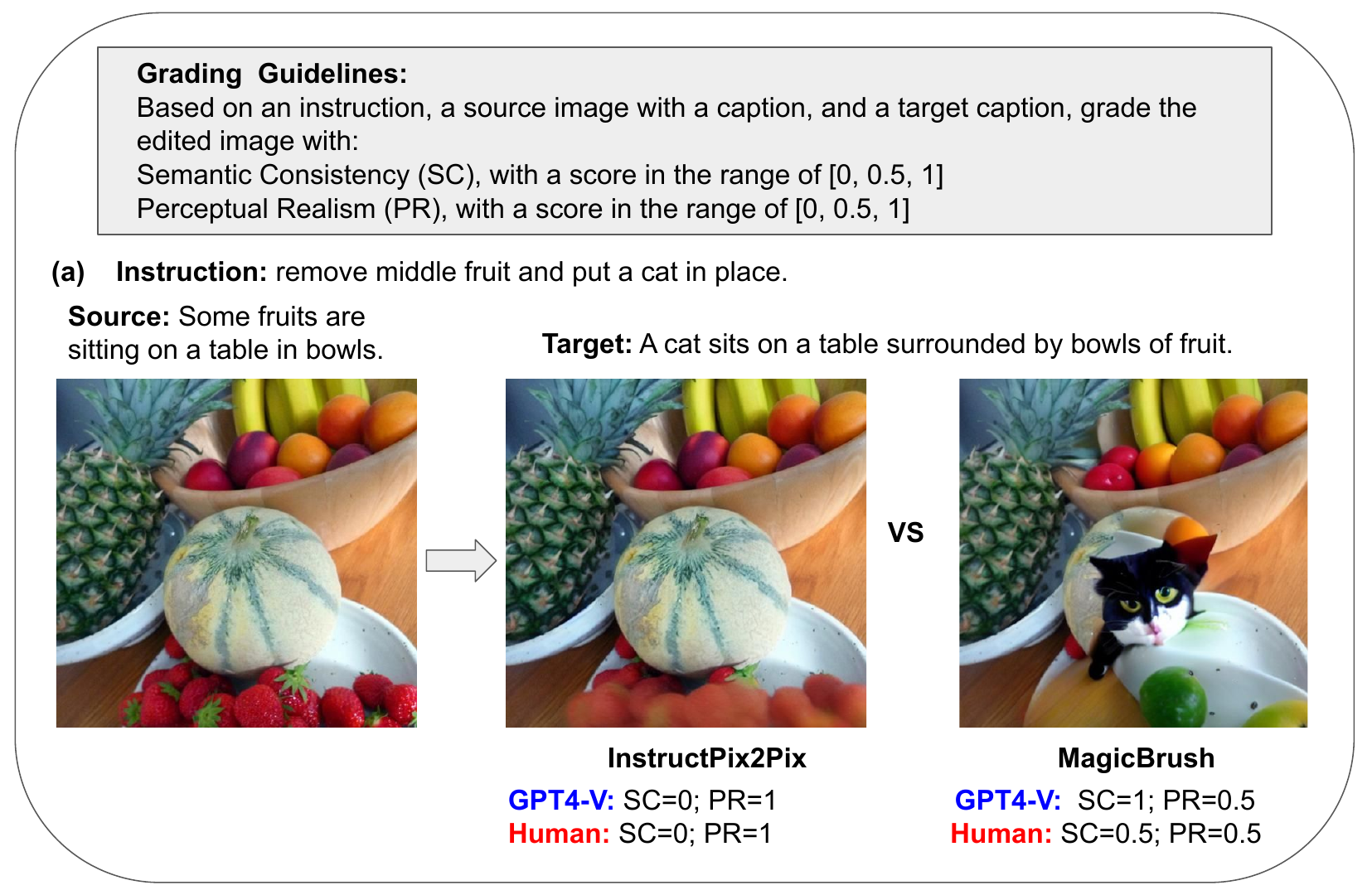} 
\caption{\textbf{Showcase of GPT-4V for evaluation text-guided image editing.} The Evaluation Guideline is following the one in ImagenHub~\citep{ku2023imagenhub}.
}
\label{fig:tgie_showcase}
\end{figure*}
Recent advances~\citep{brooks2023instructpix2pix,zhang2023magicbrush} in text-guided image editing has shown incredible results in faithfully following free-form editing instruction.
In this task, the models are required to follow instructions to edit images, where the input is a source image and an editing prompt, and the output is the target image.

\paragraph{Data Construction.} We randomly select 30 samples from ImagenHub~\citep{ku2023imagenhub} dataset, which include diversified instances with edited images synthesized by InstructPix2Pix~\citep{brooks2023instructpix2pix} and MagicBrush~\citep{zhang2023magicbrush}, etc.

\paragraph{Evaluation Setting}
The key to a successful editing requires both the correct editing of the image and keeping the background that is not requested to be changed the same.
This usually means the requirements include: 1) the synthesized image is faithful in the background compared with the source image, 2) the synthesized image follows the edit instruction and align with the target caption prompt if given any, 3) the synthesized image should be perceptually natural.
Since multiple requirements co-exist in the image editing task, the automatic evaluation is even more challenging than text-to-image synthesis evaluation.
Recent work~\citep{ku2023imagenhub} brings the evaluation in the same pipeline. We follow the evaluation principles to mainly validate two popular and state-of-the-art models, InstructPix2Pix~\citep{brooks2023instructpix2pix} and MagicBrush~\citep{zhang2023magicbrush}.
We feed the source image and edited image, together with source prompt, editing instruction and target prompt into GPT-4V, and feed the same evaluation guidelines that are provided to human evaluators.

\paragraph{GPT-4V reveals promising alignment with human evaluators in text-to-image tasks. } Following previous standardized evaluation pipeline~\citep{ku2023imagenhub}, we mainly focus on two aspects to evaluate text-guided image editing: 1) Semantic Consistency and 2) Perceptual Realism. We report the Pearson and Spearman correlation between GPT-4V and human evaluators in Table~\ref{tab:editing}.

We observe that GPT-4V is generally correlated with human evaluators in terms of semantic consistency. However it sometimes fails to well capture the perceptual realism.
And we also find that GPT-4V is giving the same preference to MagicBrush over InstructPix2Pix in terms of average Semantic Consistency and average Perceptual Realism as human evaluators.
For comparison with the existing metrics under the same evaluation guidelines, we refer to numbers reported in ImagenHub~\citep{ku2023imagenhub}, where the metrics DINO, CLIP-I get Spearman's correlation of $0.0439$, $0.0449$ and $0.0765$ for Perceptual Realism respectively, and CLIPScore gets Spearman's correlation of $0.0437$ for Semantic Consistency.
This indicates that GPT-4V as a generalist evaluator can surpass specialized metrics for this task.
We showcase one example of the evaluation in Figure~\ref{fig:tgie_showcase} comparing the GPT-4V and Human scoring, showing incredible evaluation capability of GPT-4V in this complicated setting.

\begin{table}[t]
\centering
\caption{\textbf{Comparison of Pearson and Spearman's coefficients between GPT-4V and Human across different tasks and benchmarks.} $^{*}$ indicates significant correlations (p < 0.05). Notice that the results of CLIPScore, DINO are reported in ImagenHub~\citep{ku2023imagenhub}.}

\label{tab:correlations}
\resizebox{.9\textwidth}{!}{%
\begin{tabular}{lccccc}
\toprule
  & \multicolumn{2}{c}{Semantic Consistency}  & \multicolumn{3}{c}{Perceptual Realism} \\
 \cmidrule{2-3}\cmidrule{4-6}
 & CLIPScore & GPT-4V & DINO & CLIP-I & GPT-4V  \\
\midrule
Pearson's Correlation & - & 0.4255$^{*}$ & - & - & 0.3915$^{*}$ \\
Spearman's Correlation & $0.0765$ & 0.4450$^{*}$ & $0.0449$ & $0.0765$ & 0.4090$^{*}$ \\
\bottomrule
\end{tabular}
}
\label{tab:editing}
\end{table}

%% file: subsection/multi-image.tex
One advancement of recent vision-language models~\citep{gpt4vcontribution,alayrac2022flamingo,ye2023mplug} is their ability to take interleaved image-text inputs, i.e., images can appear in any position of the input sequence. Here we consider a special case with multiple images to text generation. GPT-4V is asked to take inputs of multiple images and the text description, and evaluate the alignment between multiple images and the given text description.

\paragraph{Data Construction.} We select a subset from the Gest dataset~\citep{yan2023personalized} for multi-image to text generation. The dataset consists of user uploaded images and corresponding reviews crawled from Google Maps. Specifically, to construct the evaluation set, we randomly sample 50 user reviews which contain 2-4 images, since currently GPT-4V only supports a maximum of 4 images in a single dialogue.

\begin{figure*}[t!] 
\centering 
\includegraphics[width=0.9\textwidth]{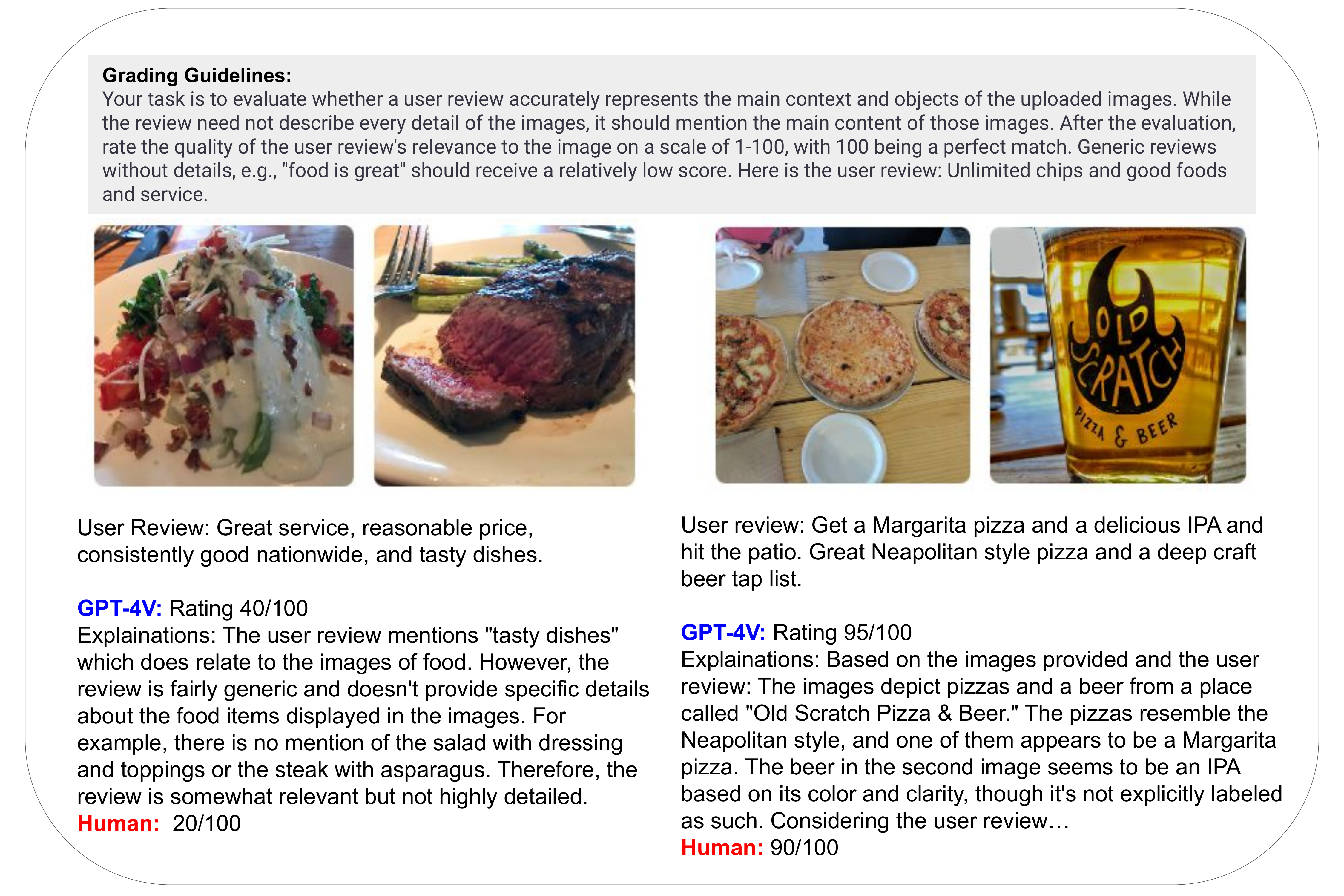} 
\caption{\textbf{Examples of GPT-4V for evaluating multiple images to text task.} GPT-4V successfully identifies the main objects of each image and provides comprehensive explanations for its scores.
}
\label{fig:i2t_google_fig}
\end{figure*}

\paragraph{Evaluation Setup.} Following previous evaluation setting~\citep{yan2023personalized}, in this task, we evaluate the alignment between the multiple images and the user review.  Here our scoring system is GPT-4V-Single, where GPT-4V is instructed to generate a score to evaluate the quality and alignment of the multiple-image input and text output pair.
For example, a high score is given if the user review mentions the dishes in the images. A generic review such as ``great food and service'' without mentioning the detailed content of those images will receive low scores.  Both human and GPT-4V are asked to grade each sample based on the visual alignment between the review and the images, on a scale of 1-100.  Two examples of input images and user reviews, along with GPT-4V and human scores are presented in Figure~\ref{fig:i2t_google_fig}. 
As a generalist evaluator, GPT-4V need to accurately under the internal-relation between multiple images and the relation between images and text.

\paragraph{GPT-4V has the potential of being a strong evaluator for multi-image to text evaluation.}
Pearson and Spearman Correlations between GPT-4V's scoring and human scoring are computed to evaluate GPT-4V's performance for this task.

Shown in Table~\ref{tab:i2t_google_table},  both Pearson and Spearman correlation scores are around 0.8, suggesting that there is a strong positive relationship between the GPT-4V scores and human scores. This indicates that GPT-4V successfully understands the internal-relation between multiple images and the relation between images and text, showing the potential of using GPT-4V as a evaluator for evaluating task.
Additionally, since GPT-4V also generates its reasoning process in text by thinking step by step~\cite{kojima2022large}, it provides an appealing property for interpret model decisions. We manually evaluate the generated text descriptions of each image from GPT-4V, we find that GPT-4V is good at accurately describing each image separately, outperforming our observations on recent open-source VLMs~\citep{ye2023mplug}, which often fail to capture the correct information of each image in a input sequence.

\begin{table}[h]
\centering
\vspace{-1em}
\caption{\textbf{Average evaluation scores from GPT-4V and human, along with Pearson and Spearman correlations between the two.} Results on evaluated on the Google Local dataset. $^{*}$ indicates significant correlations (p < 0.05).
}
\begin{tabular}[t]{l c c c c}
\toprule
 & GPT-4V-AVG & Human-AVG & Pearson & Spearman\\
\midrule
 & 68.90 & 60.10  & 0.826$^{*}$ &  0.794$^{*}$ \\
 \bottomrule
\end{tabular}
 \label{tab:i2t_google_table}
\end{table}

%% file: subsection/relatedwork.tex
\paragraph{Multimodal Large Language Models.} Driven by the success of text-only Large Language Models (LLMs), i.e. GPT-4~\citep{gpt4}, Claude~\citep{claude}, Llama~\citep{llama}, Multimodal Large Language Models (MLLMs) are the next step towards Artificial General Intelligence (AGI) and arouses high research attentions. There are mainly two research lines in developing strong MLLMs, one-stage multimodal pre-trained models and two-stage vision-language alignment models. The one-stage pre-trained MLLMs like Fuyu~\citep{fuyu-8b} and Kosmos~\citep{kosmos} are directly pre-trained on mixed large-scale multimodal corpus, including text-only corpora, image-text paired data, and image-text interleaved data, which leads to strong generalization capability. Specifically, Fuyu processes the image as patches for embedding, and Kosmos adopts CLIP~\citep{clip} for encoding image emnbedding. Another type of MLLMs are constructed on the strong pre-trained instruction following LLMs and vision encoders. MiniGPT-4~\citep{zhu2023minigpt} adopts Q-former of pre-trained BLIP-2~\citep{blip2} as image encoder and Vicuna~\citep{vicuna2023} as the text generator, while LLaVA~\citep{liu2023llava} adopts CLIP as the image encoder for simplicity. The two-stage vision-language alignment involves the feature alignment stage to unify the embedding space of image and text, and the multimodal instruction tuning stage to engage the model to follow multi-modal instructions.

\paragraph{Large Language Model as Interpretable Reference-free Metric.} With the strong zero-shot task completion capability, LLMs are deployed as the replacement of human evaluator and automatic evaluator on open-ended language generations tasks. G-Eval~\citep{geval} investigates the robustness and trustfulness of adopting GPT-4 to evaluate the quality of summaries generated by models in an single-summary grading manner. LLM-as-a-judge~\citep{llmjudge} adopts LLMs to evaluate the task completion answer quality for LLMs in settings of single-sample grading, pairwise comparison, and reference-guided grading.
In multi-modal domain, reference-based automatic evaluation metrics like Rouge~\citep{lin-2004-rouge}, and METEOR~\citep{banerjee-lavie-2005-meteor} are usually constrained by the lack of references in some cases, such as text-to-image synthesis. Then recently proposed reference-free metrics ~\citep{hessel-etal-2021-clipscore, blip2, xu2023imagereward, wu2023human, kirstain2023pickapic} are able to capture the text-image alignment. More recent metrics~\citep{hu2023tifa,lu2023llmscore} show promising results by utilizing large pre-trained models to enable the interpretable evaluations for vision-language tasks.
Nevertheless, none of these metrics are capable of adapting to diversified evaluation guidelines and suit for various tasks. In this work, we explore the potential of GPT-4V as a universal reference-free metric that align with human preference in a broad domain of vision-language tasks.

%% file: subsection/limitation.tex
\begin{figure*}[t!] 
\centering 
\includegraphics[width=0.9\textwidth]{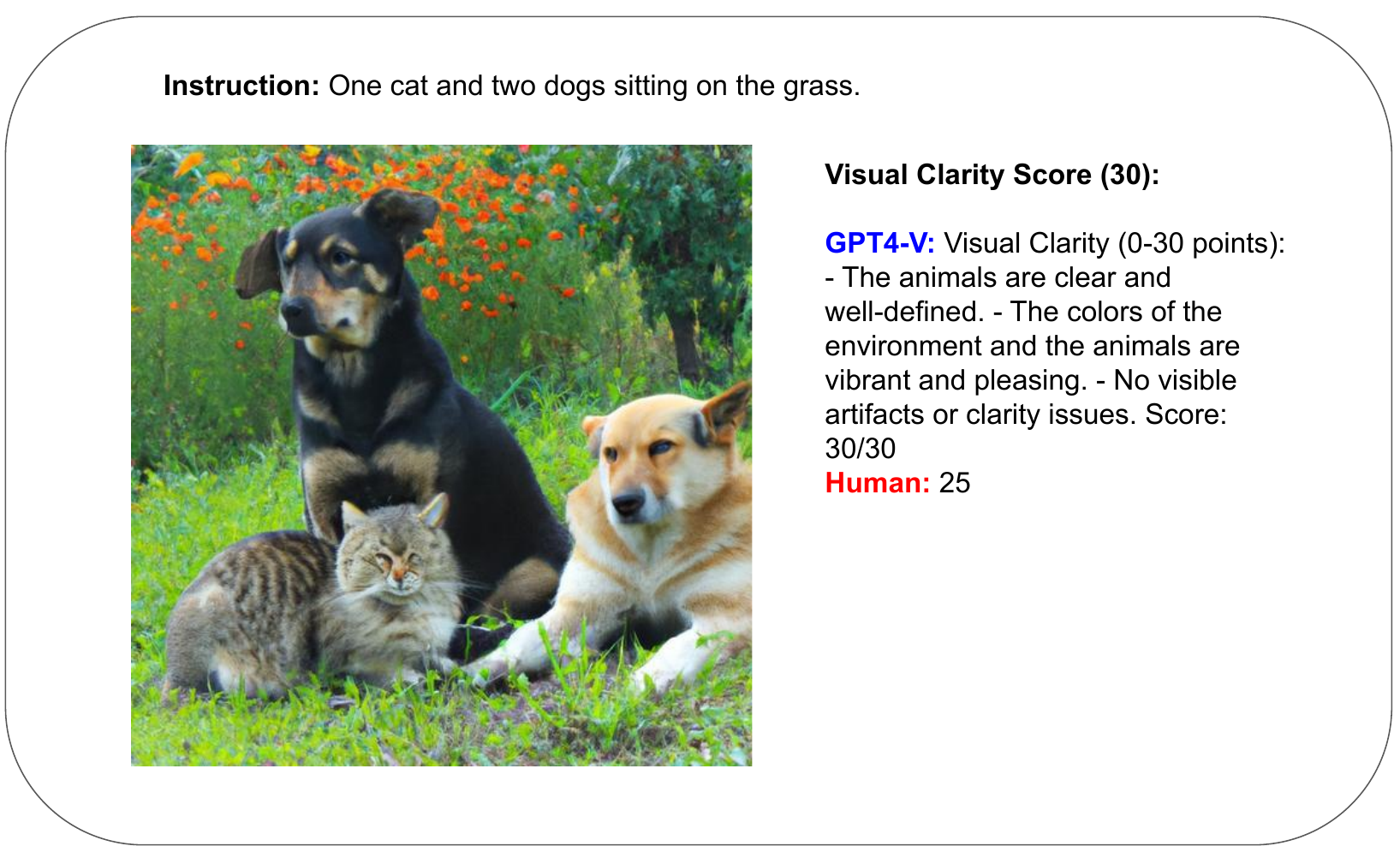} 
\caption{\textbf{GPT-4V's Visual Clarity Scoring: A Comparative Analysis.} While GPT-4V awarded a perfect score for visual clarity, human evaluators marked the image down due to the unclearness of the cat's and the yellow dog's faces.}
\label{fig:vs_score1}
\end{figure*}

\begin{figure*}[t!] 
\centering 
\includegraphics[width=0.9\textwidth]{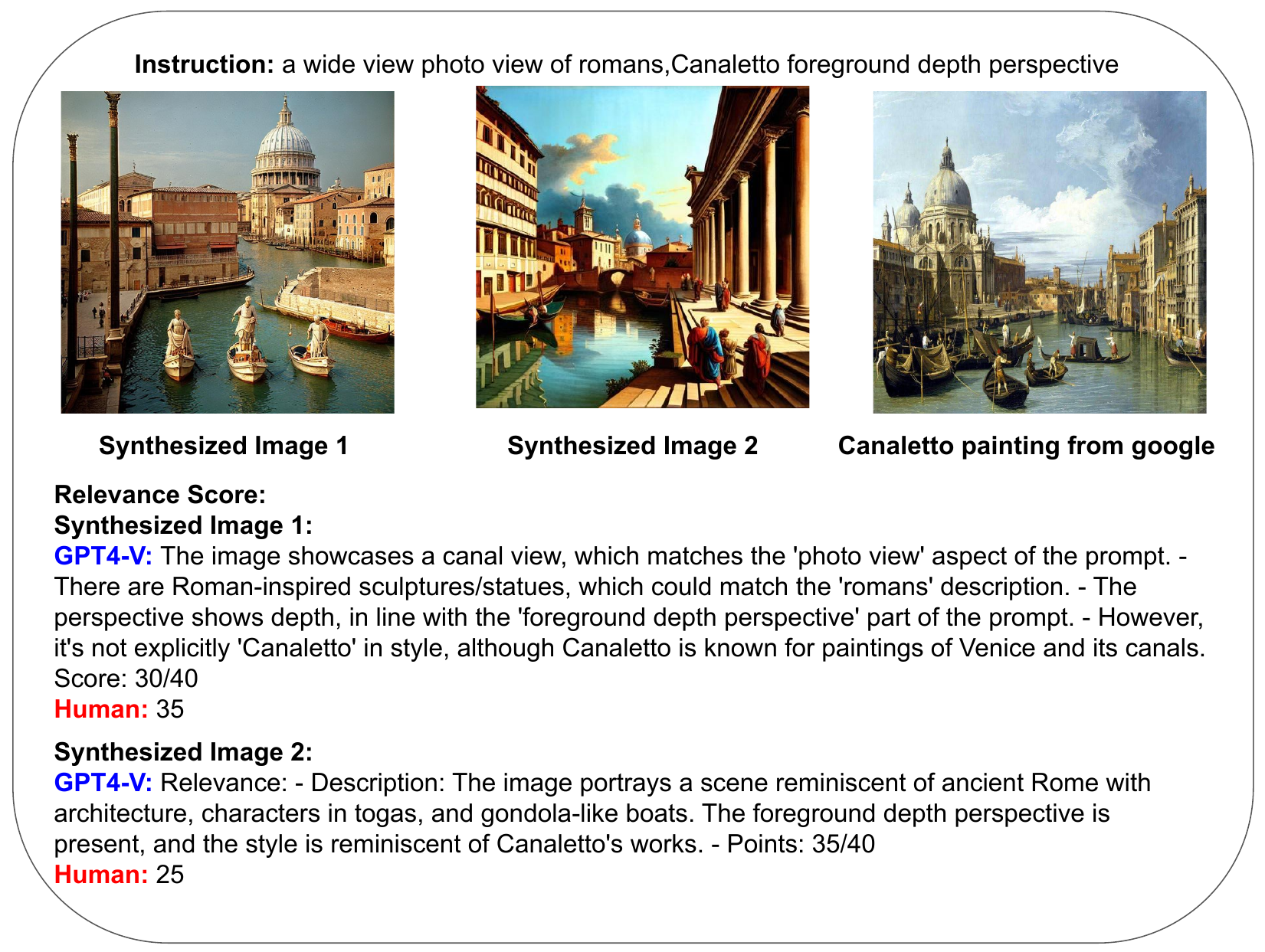} 
\caption{\textbf{GPT-4V's Knowledge-Intensive Task Evaluation.} Comparison of GPT-4V's evaluation on images with Canaletto's painting style, highlighting challenges in knowledge-intensive tasks.}
\label{fig:vs_score2}
\end{figure*}

\paragraph{Limited capacity in grading image with complex caption.}
We discovered that GPT-4V performs significantly better in image-to-text tasks compared to text-to-image tasks. This discrepancy could be attributed to the complexity and variability of real-world captions used for generating images. For instance, in Figure \ref{fig:t2i_all} (b), the caption utilized for image generation is long and cluttered, containing a mix of both abstract and detailed requirements. In contrast, the outputs for image-to-text tasks are easier to understand, more generalized, and well-formatted, as demonstrated in Figure  \ref{fig:i2t_exp}.

\paragraph{Discrepancies in Visual Clarity Assessments}
Our examination revealed that GPT-4V exhibits potential challenges in discerning subtle differences in visual clarity.
This is illustrated in Figures \ref{fig:vs_score1}, where GPT-4V attributed higher scores for visual clarity compared to those determined by human evaluators. Human assessors allocated a score of 25 out of a maximum of 30, noting that the visual indistinctness of facial features in both the cat and the yellow dog detracted from the overall clarity of the images. This suggests that GPT-4V might struggle with capturing and evaluating fine details in visual quality, resulting in a comparatively reduced effectiveness in text-to-image evaluation tasks. In contrast, the image-to-text tasks appear to be less affected by these clarity issues, as they do not emphasize the sharpness of image details as heavily.

\paragraph{Challenges with Knowledge-Intensive Tasks.}
GPT-4V appears to encounter difficulties when tasked with evaluations that demand specific expertise. As illustrated in Figure~\ref{fig:vs_score2}, the evaluation required an understanding of Canaletto's painting style. Surprisingly, GPT-4V awarded a higher score to Synthesized Image 2 than Synthesized Image 1, even though Synthesized Image 1 more closely mirrors Canaletto's style, as can be observed from a reference Canaletto painting sourced via a Google search.

\begin{table*}[t!]
	\centering 
 \caption{ \textbf{Position bias analysis of different tasks for GPT-4V evaluation. }Consistency(\%) is a
evaluator gives consistent results when swapping the order of two assistants. “Biased toward first/second” (\%) is a evaluator favors the first/second answer. }
 \vspace{-2mm}
 \renewcommand{\arraystretch}{0.95}

\begin{tabular}{l c cc cc }
\toprule
task & \# instances & consistency & biased toward first & biased toward second \\ 
\midrule
image-to-text & 100 & 92\% & 3\% & 2\% \\
text-to-image & 70 & 51.5\% & 10.1\% & 9.1\%  \\
\bottomrule
\end{tabular}
\label{tab:postion}
\end{table*}

\paragraph{Position Bias Analysis.}
Position bias is pointed out in previous research single-modal evaluator~\citep{llmjudge}. In our primary study, minimal position bias was noted in the image-to-text and text-to-image tasks, as shown in \ref{tab:postion}. We offer the following explanations:
\begin{itemize}
    \item For the image-to-text task, GPT-4V efficiently handled hard negative captions introduced with subtle noise, thereby reducing the potential for pronounced position bias.

    \item As for the text-to-image task, a nearly equal bias was observed toward both the first and second positions. This balanced bias might be a result of GPT-4V's underlying multimodal architecture. To gain a clearer understanding, we intend to conduct more comprehensive experiments with an expanded sample size.  
\end{itemize}

%% file: subsection/app.tex
\section{Image-to-Text Captioning}
\subsection{Prompt}
\begin{tcolorbox}[enhanced]
\textbf{Single-Answer Grading: }\\
Your task is to evaluate whether a given text caption accurately represents the main content and objects of an associated image. While the caption need not describe every detail of the image, it should convey the overall theme or subject. After your evaluation, rate the quality of the text caption's match to the image on a scale of 1-100, with 100 being a perfect match.\\
\\
\textbf{Pairwise Comparison : }\\
Select between caption 0 and caption 1, according to which one you believe aligns most accurately with the provided image. In cases where both captions seem to possess equal quality in adherence to the image, respond with 'Tie'. Your selection can be subjective. Give your intermediate thinking step by step before giving the final response. Your final response must be 'caption 0', 'caption 1', or 'Tie'. Output your final answer in the format ""Final Answer: caption 0/caption 1/Tie.
\end{tcolorbox}

\subsection{More Showcases}

\begin{figure*}[ht] 
\centering 
\includegraphics[width=\textwidth]{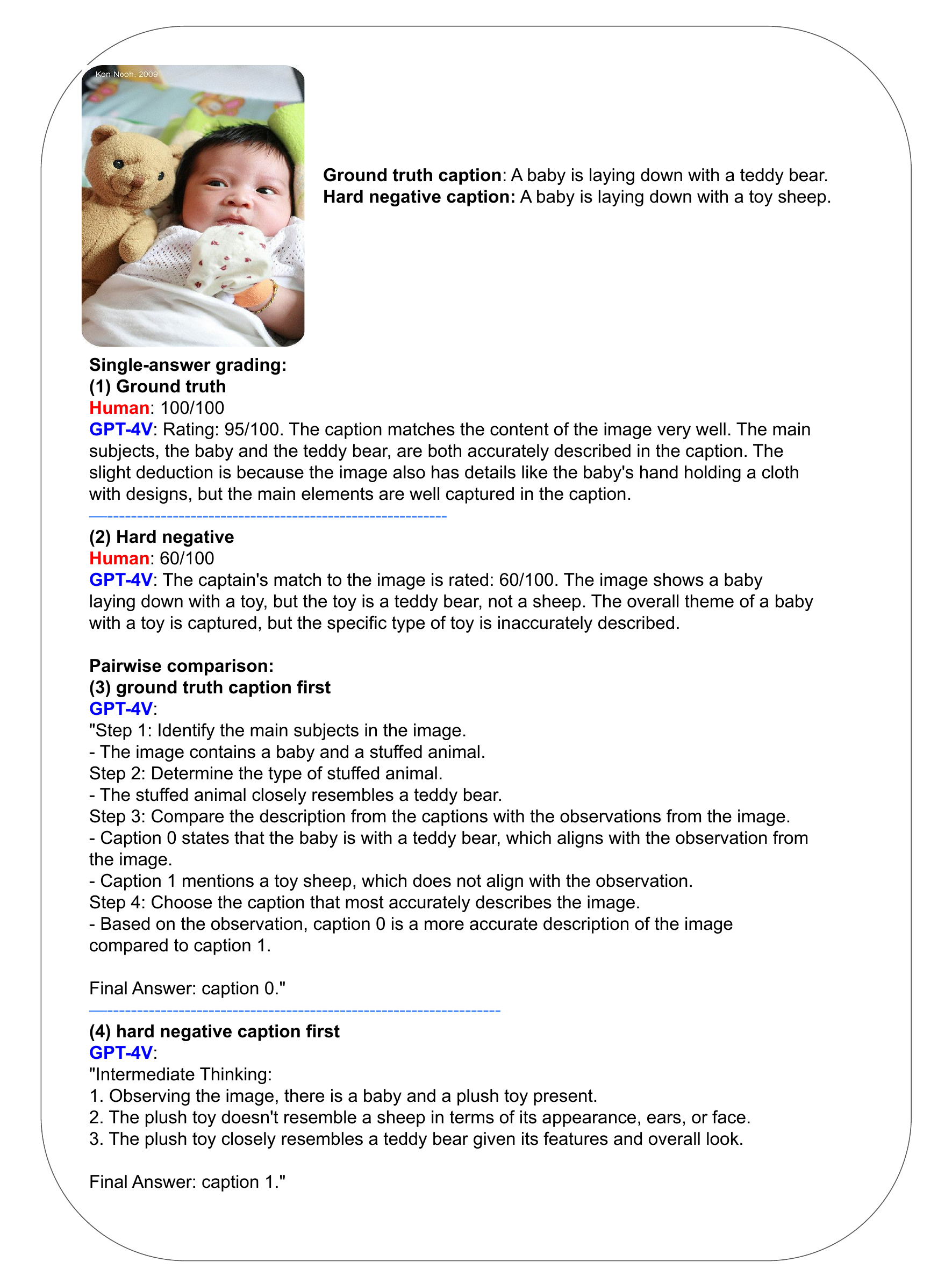} 
\caption{Example of grading on image-to-text evaluation. 
}
\label{fig:i2t_example_1}
\end{figure*}

\begin{figure*}[ht] 
\centering 
\includegraphics[width=\textwidth]{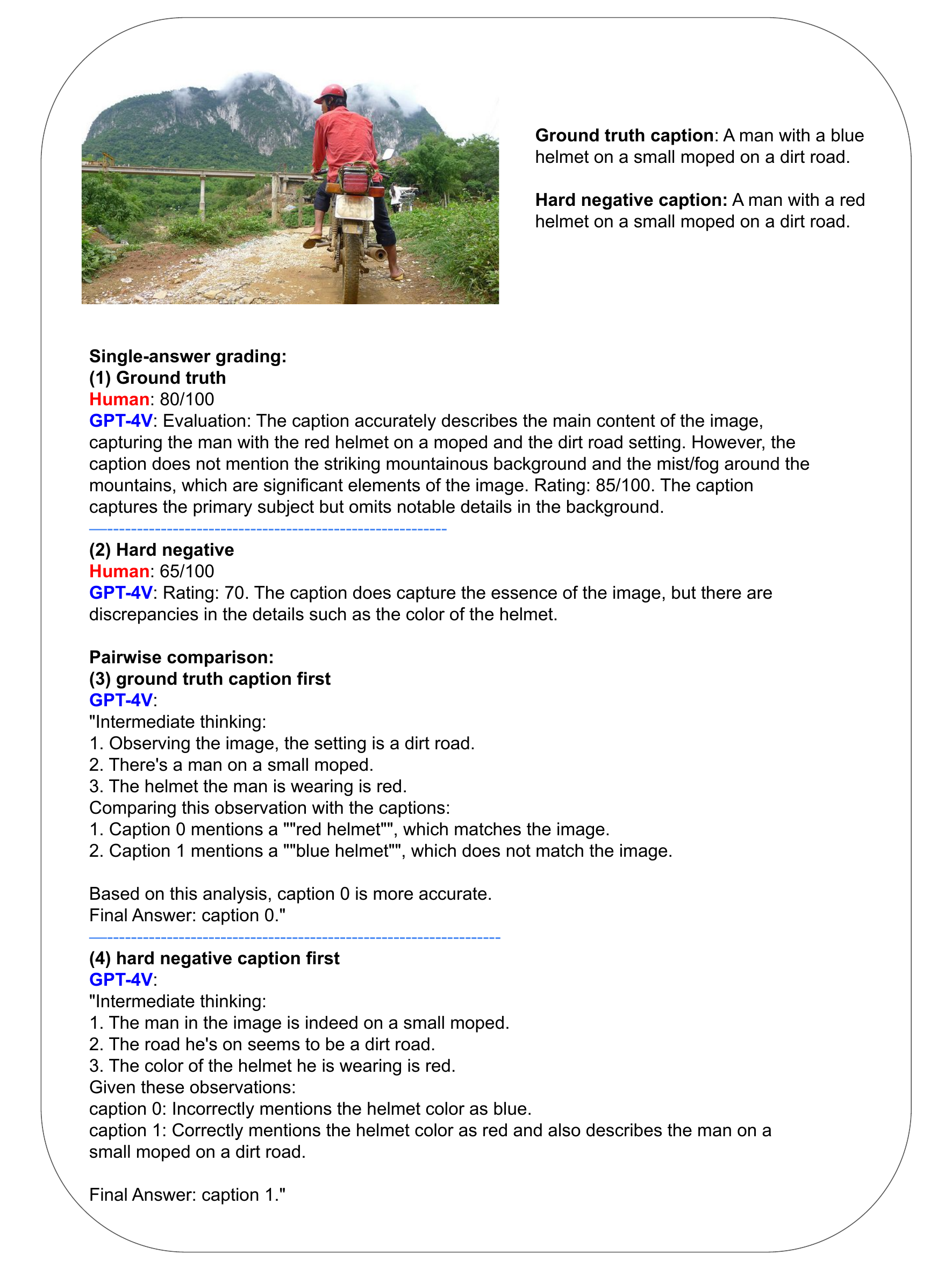} 
\caption{Example of grading on image-to-text evaluation. 
}
\label{fig:i2t_example_2}
\end{figure*}

\begin{figure*}[ht] 
\centering 
\includegraphics[width=\textwidth]{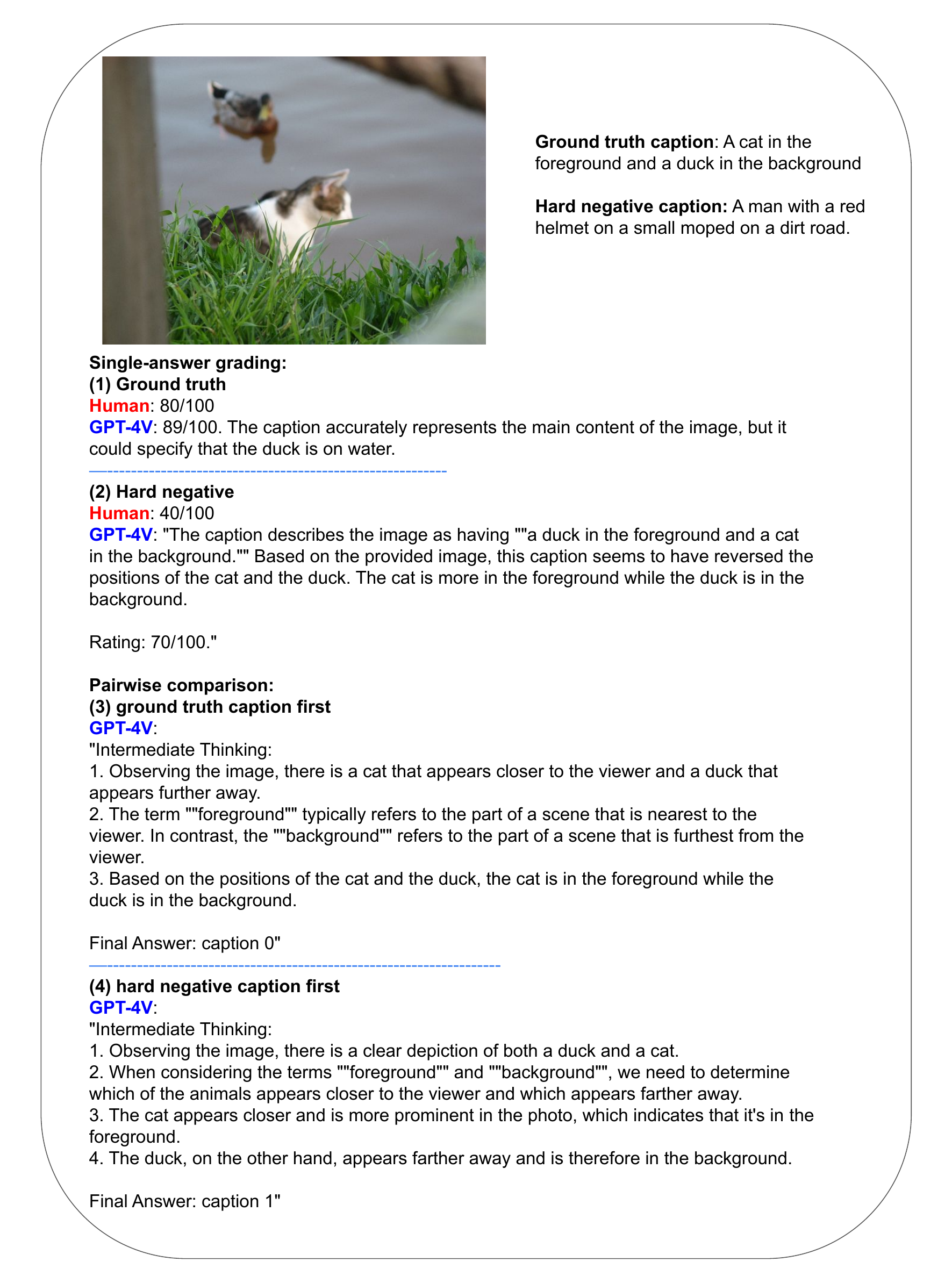} 
\caption{Example of grading on image-to-text evaluation. 
}
\label{fig:i2t_example_3}
\end{figure*}

\clearpage
\newpage
\section{Text-to-Image Generation }
\subsection{Prompt}
\begin{tcolorbox}
\textbf{Single-Answer Grading: }\\
Prompt Given to Image Generation Model: [[[\textit{The instruction to image generation }]]]\\
\\
Instructions: Carefully assess the generated image in terms of relevance to the prompt, visual clarity, and object accuracy. Use the following criteria to guide your evaluation:\\
\\
Relevance (0-40 points):\\
How closely does the generated image match the prompt?\\
Does the image capture the core essence and details specified in the prompt?\\
\\
Visual Clarity (0-30 points):\\
How clear and sharp is the image?\\
Are there any artifacts or oddities that detract from the image's quality?\\
Are the colors and contrasts pleasing and coherent?\\
\\
Object Accuracy (0-30 points):\\
Are all the nouns from the prompt represented accurately in the image?\\
Do the depicted objects match the description and intention of the original nouns in the prompt?\\
Is there any object that doesn't belong or was not mentioned in the prompt?\\
\\
Total Score: \_\_/100\\
\\
\textbf{Pairwise Comparison : }\\
Select between image 0 (located on the left) and image 1 (located on the right), according to which one you believe aligns most accurately with the provided instructions. In cases where both images seem to possess equal quality in adherence to the instructions, respond with 'Tie'.\\
\\
Your selection can be subjective. Give your intermediate thinking step by step before give the final respond.  Your final response must be 'image 0', 'image 1' or 'Tie'.  Output your final answer as the format "Final Answer: image 0/image 1/Tie."\\
\\
Instruction:[[[\textit{The instruction to image generation }]]]
\end{tcolorbox}

\subsection{More Showcases}

\begin{figure*}[ht] 
\centering 
\includegraphics[width=\textwidth]{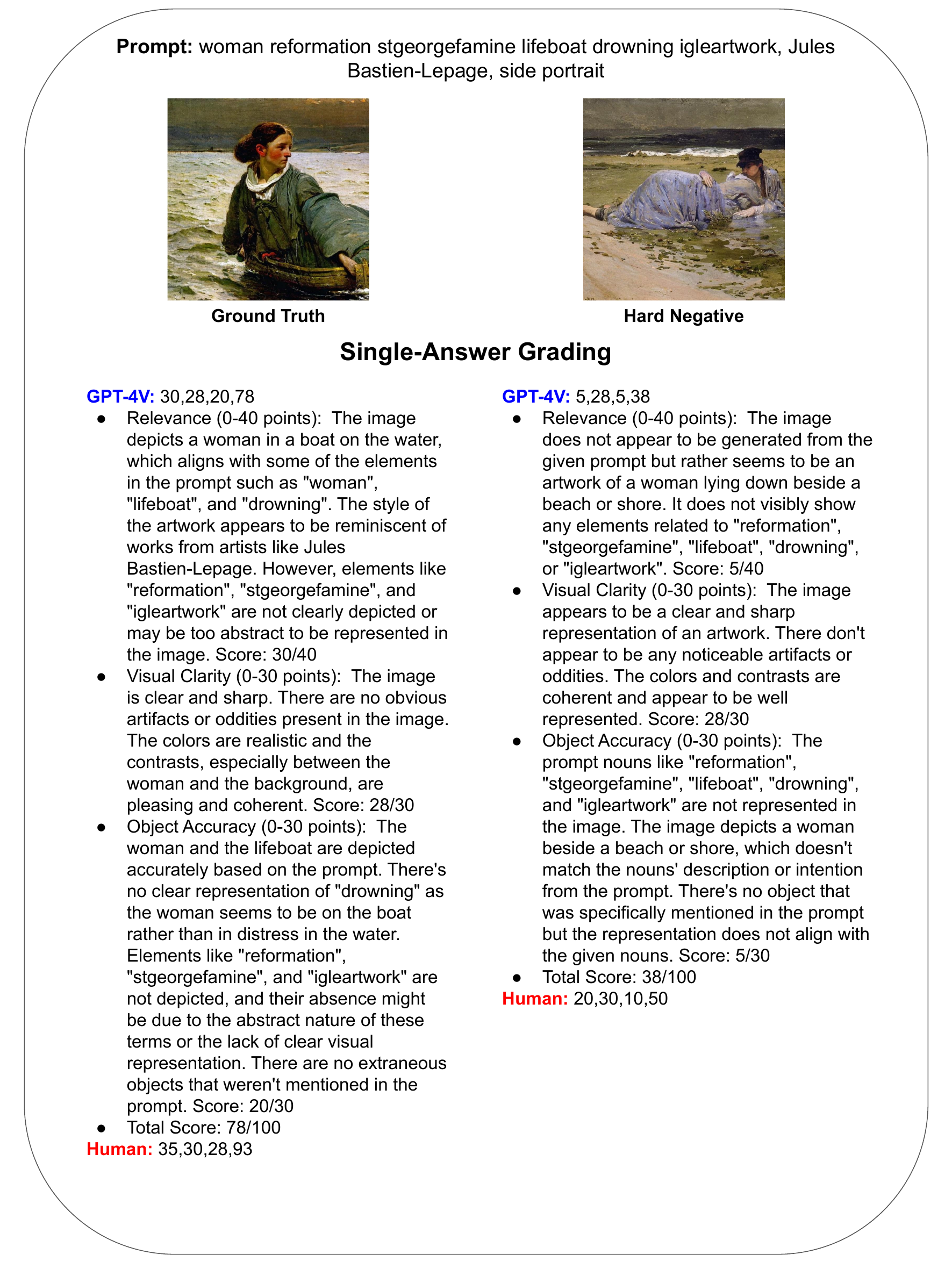} 
\caption{Example of single-answer grading on text-to-image evaluation. 
}
\label{fig:t2i_example_1_single}
\end{figure*}

\begin{figure*}[ht] 
\centering 
\includegraphics[width=\textwidth]{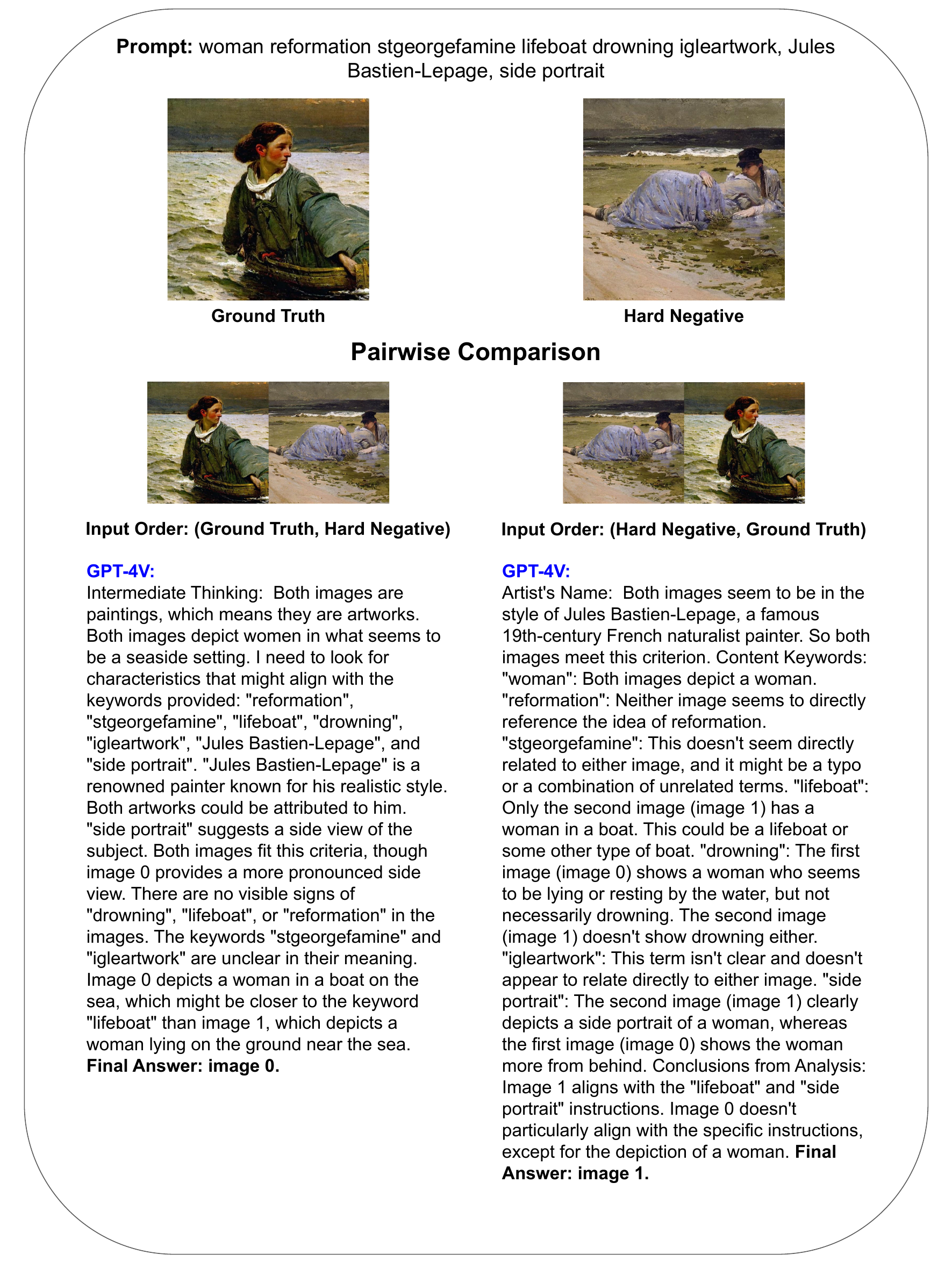} 
\caption{Example of pairwise comparison on text-to-image evaluation. 
}
\label{fig:t2i_example_1_pairwise}
\end{figure*}

\begin{figure*}[ht] 
\centering 
\includegraphics[width=\textwidth]{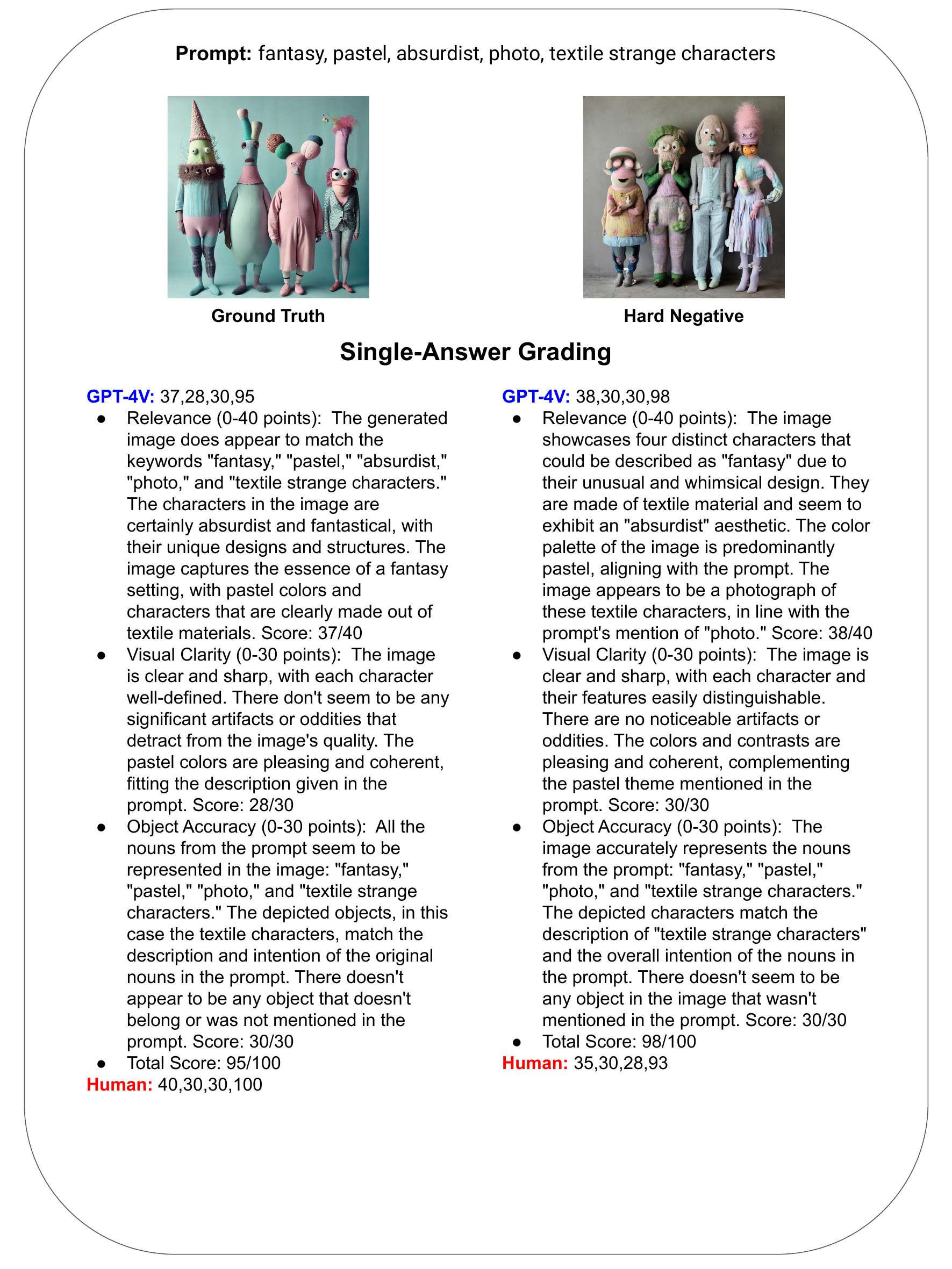} 
\caption{Example of single-answer grading on text-to-image evaluation. 
}
\label{fig:t2i_example_2_single}
\end{figure*}

\begin{figure*}[ht] 
\centering 
\includegraphics[width=\textwidth]{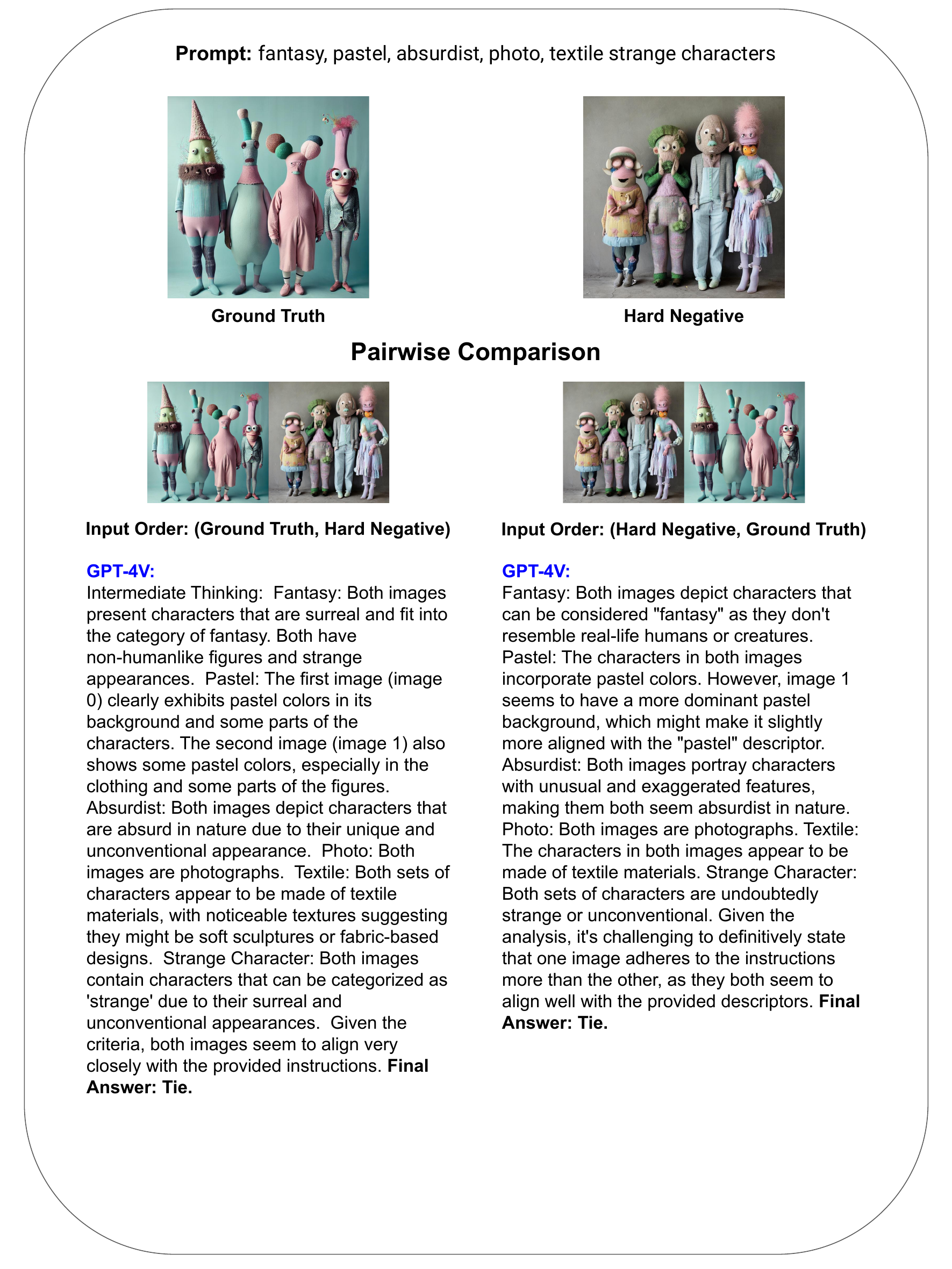} 
\caption{Example of pairwise comparison on text-to-image evaluation. 
}
\label{fig:t2i_example_2_pairwise}
\end{figure*}

\begin{figure*}[ht] 
\centering 
\includegraphics[width=\textwidth]{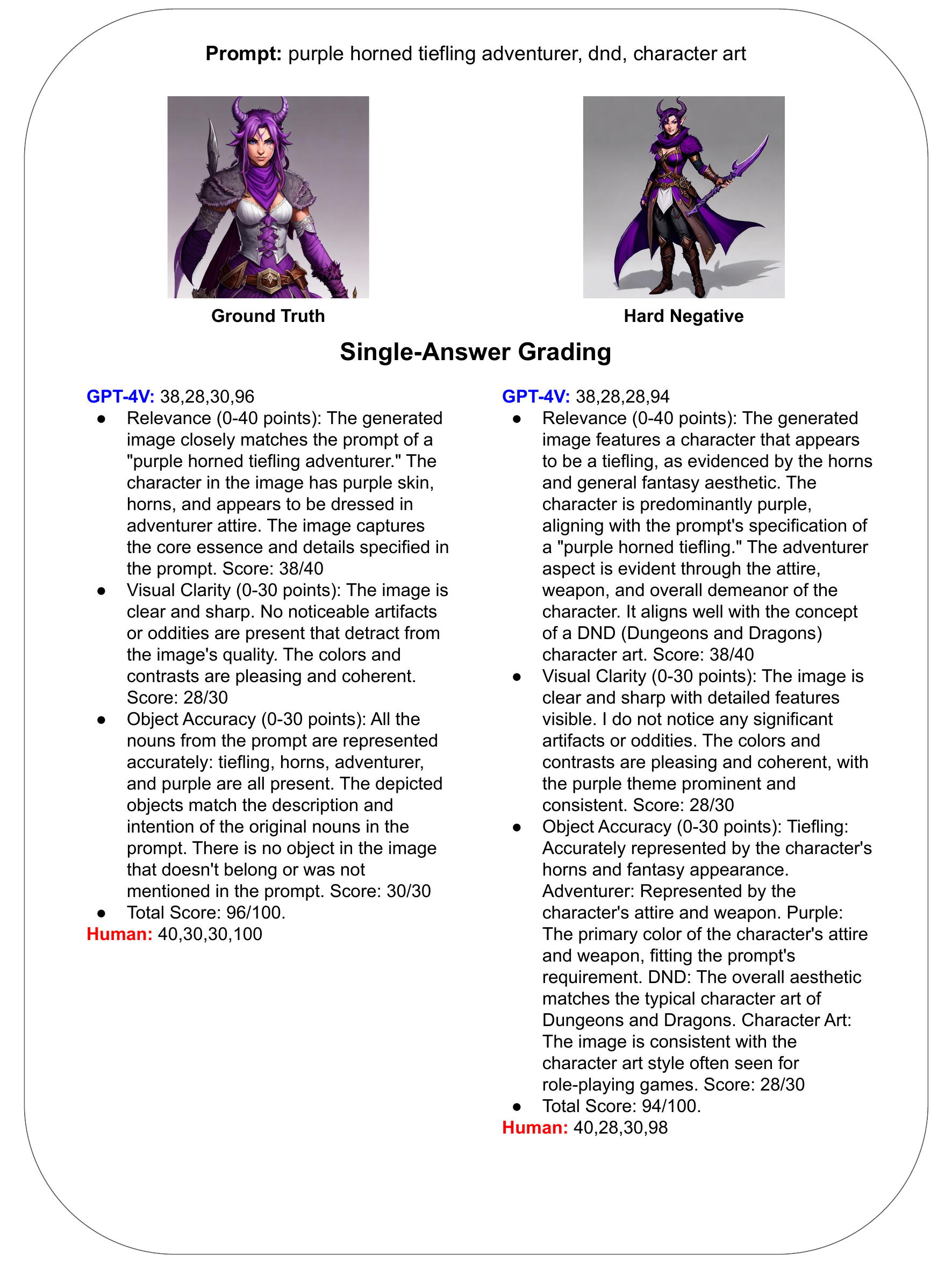} 
\caption{Example of single-answer grading on text-to-image evaluation. 
}
\label{fig:t2i_example_3_single}
\end{figure*}

\begin{figure*}[ht] 
\centering 
\includegraphics[width=\textwidth]{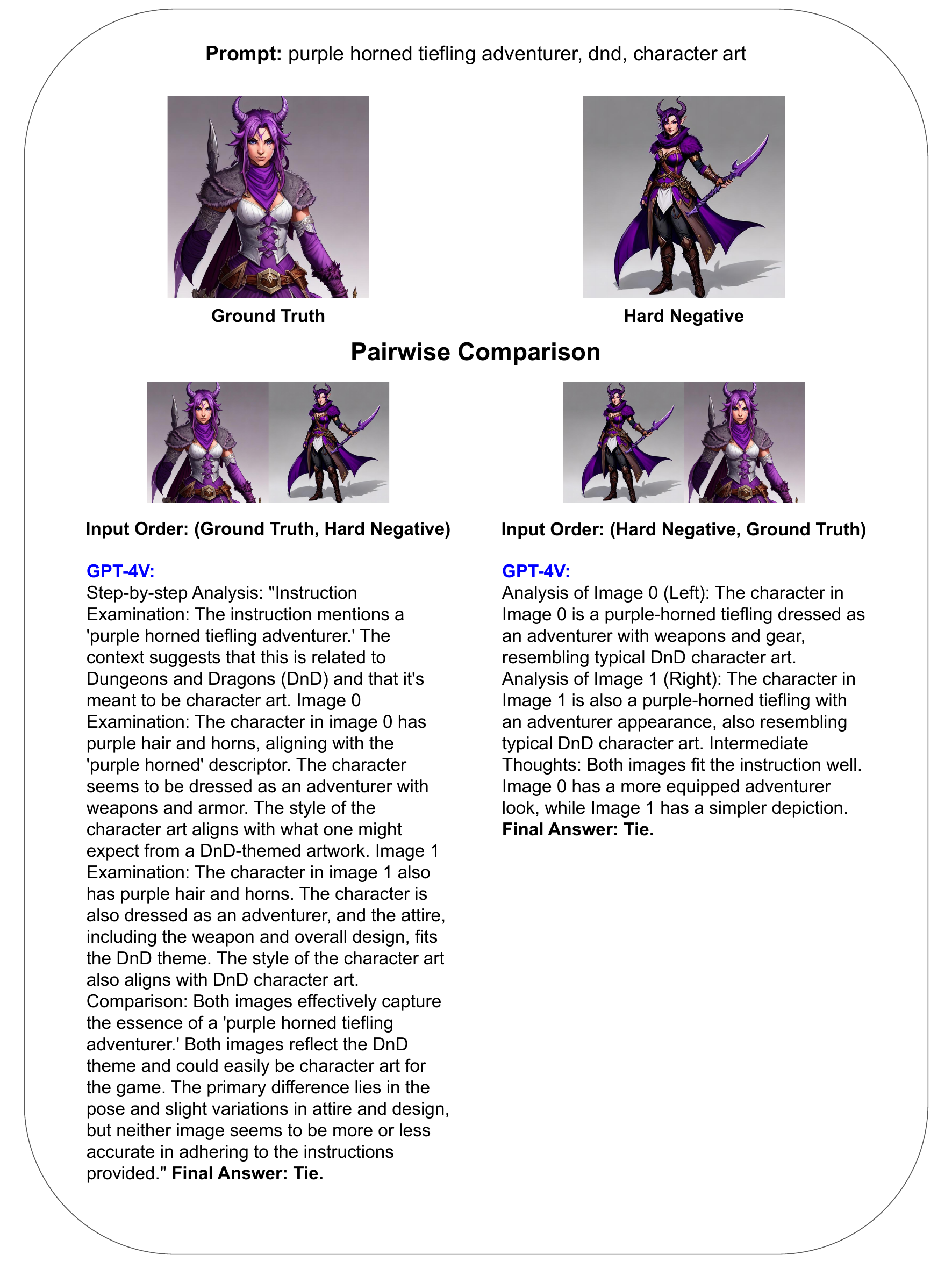} 
\caption{Example of pairwise comparison on text-to-image evaluation. 
}
\label{fig:t2i_example_3_pairwise}
\end{figure*}
\clearpage
\newpage

\section{Text-guided Image Editing }

\subsection{Prompt}
\begin{tcolorbox}
\textbf{Single-answer grading: }\\
To standardize the conduction of a rigorous human evaluation, we stipulate the criteria for each measurement as follows:\\
Semantic Consistency (SC), score in range [0, 0.5, 1]\\
Perceptual Realism (PR), score in range [0, 0.5, 1]\\
Semantic Consistency (SC) ensures that the generated image is coherent in terms of guidance provided (i.e. Prompts, Subject Token, etc.). In another words, the image has to be aligned with the requirements provided in user's inputs.\\
Perceptual Realism (PR) ensures the generated image align with real-world characteristics. In another words, the image has to be visually convincing and closely resembles a real photograph (Photorealism).\\

General Rules for Semantic Consistency (SC) scoring:\\
SC=0 : Image not following one or more of the conditions at all (e.g. not following the prompt at all, different background in editing task, wrong subject in subject-driven task etc)
SC=0.5 : all the conditions are partly following the requirements.
SC=1 : All the conditions are following >75\% of the requirement. You agree that the overall idea is correct.\\

General Rules for Perceptual Realism (PR) scoring:\\
PR=0: Obvious distortion/artifacts that are unrecognizable at first glance
PR=0.5: Some artifacts but the objects are still recognizable; or Unnatural sense of detail feeling in some area (You find out the image looks strange after examining it carefully.)
PR=1: You agree that the image generally look real (doesn't have to be 100\% perfect. Like 90\% is good enough.)\\
\end{tcolorbox}
Notice that this evaluation guideline is following the standardized one in ImagenHub~\citep{ku2023imagenhub}.
\subsection{More Showcases}
\begin{figure*}[t] 
\centering 
\includegraphics[width=0.9\textwidth]{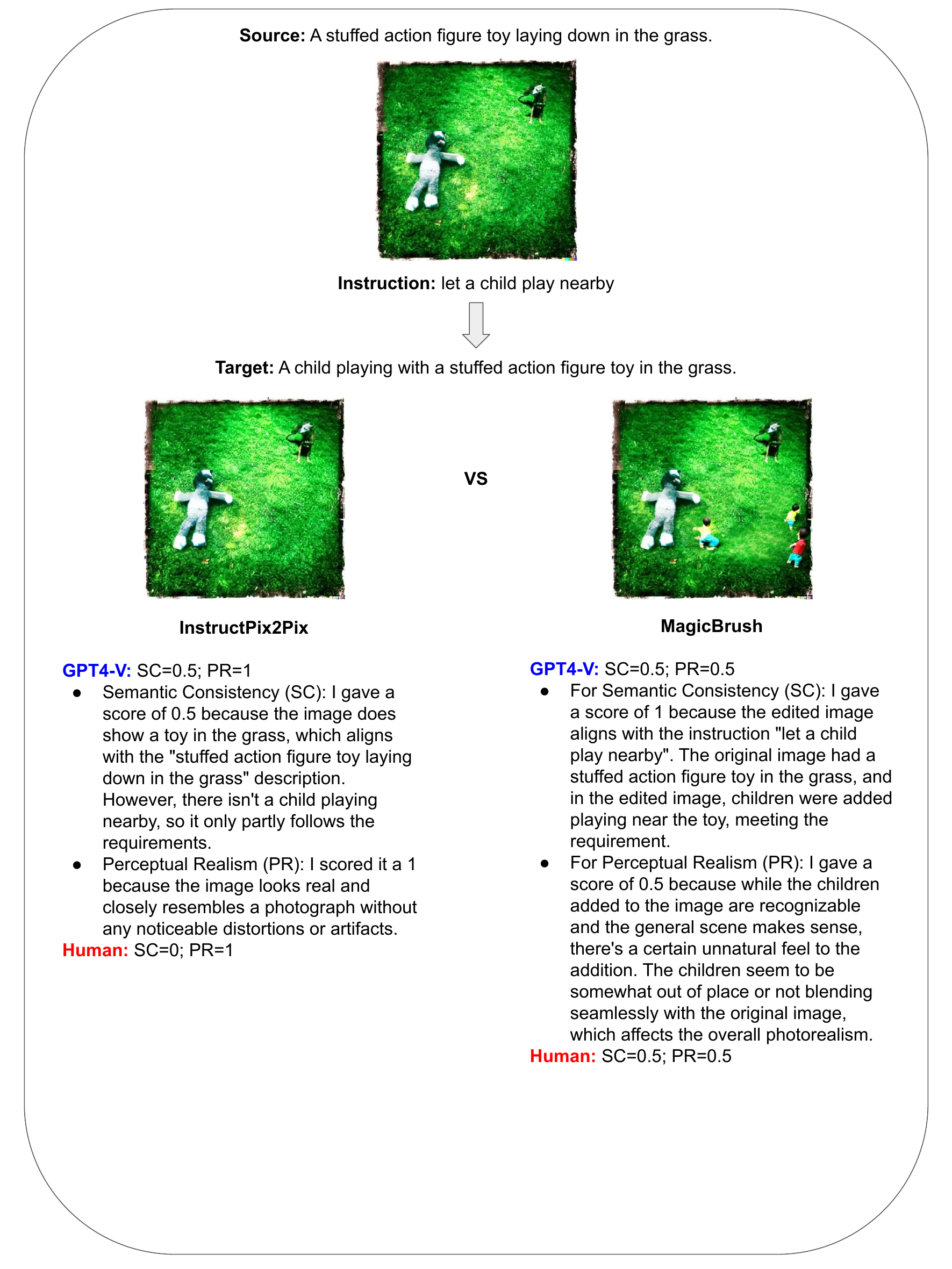} 
\caption{Showcase of GPT-4V for evaluation text-guided image editing.
}
\label{fig:tgie_showcase2}
\end{figure*}
\begin{figure*}[t] 
\centering 
\includegraphics[width=0.9\textwidth]{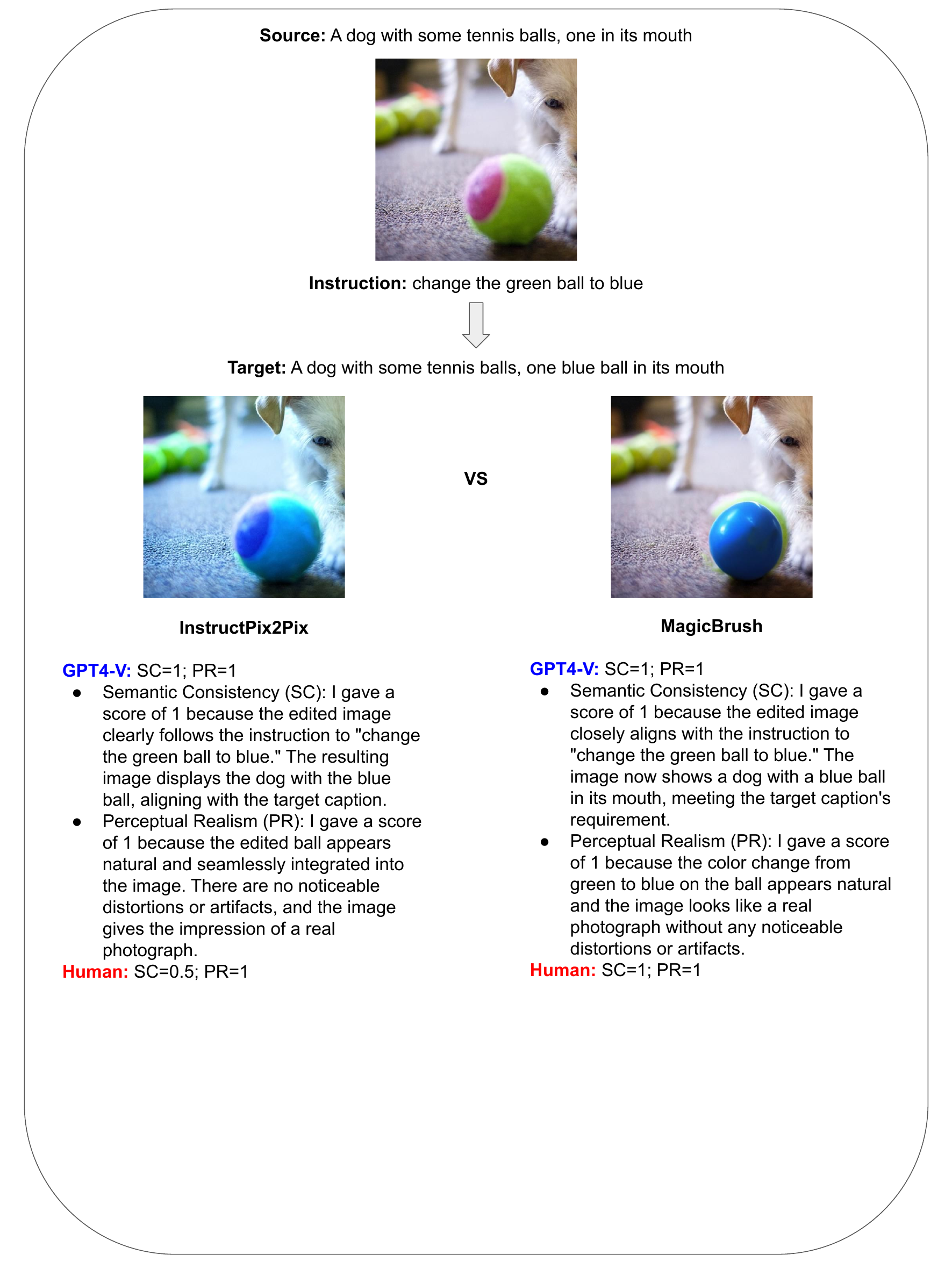} 
\caption{Showcase of GPT-4V for evaluation text-guided image editing.
}
\label{fig:tgie_showcase3}
\end{figure*}
\begin{figure*}[t] 
\centering 
\includegraphics[width=0.9\textwidth]{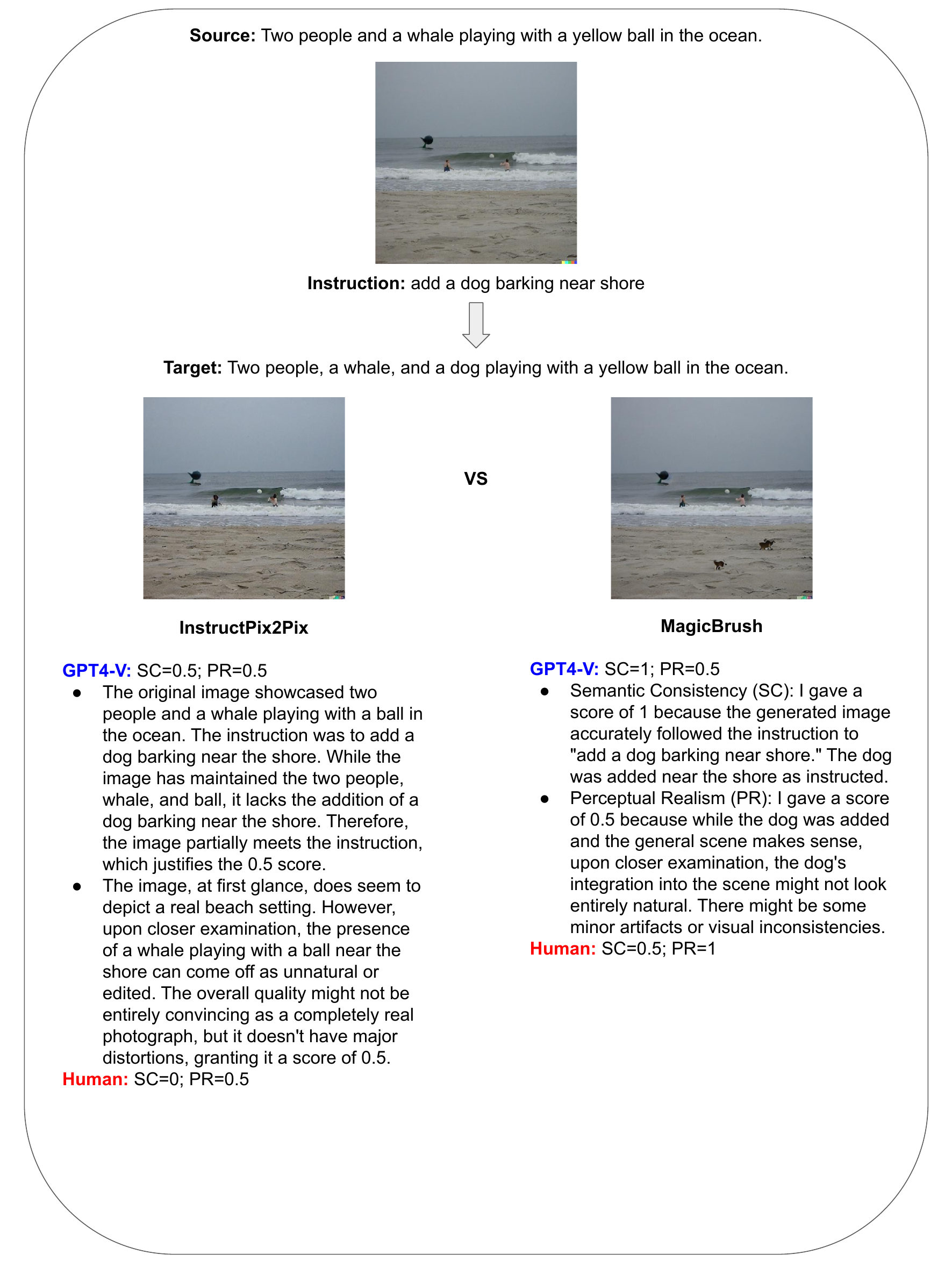} 
\caption{Showcase of GPT-4V for evaluation text-guided image editing.
}
\label{fig:tgie_showcase4}
\end{figure*}

\clearpage
\newpage
\section{Multiple Images to Text Alignment}
\begin{tcolorbox}[enhanced]
\textbf{Single-Answer grading: }\\
Your task is to evaluate whether a user review accurately represents the main context and objects of the uploaded images. While the review need not describe every detail of the images, it should mention the main content of those images. After the evaluation, rate the quality of the user review's relevance to the image on a scale of 1-100, with 100 being a perfect match. Generic reviews without details, e.g., "food is great" should receive a relatively low score. Here is the user review: [[[[\textit{text review information.}]]]\\

\end{tcolorbox}

\begin{figure*}[h] 
\centering 
\includegraphics[width=\textwidth]{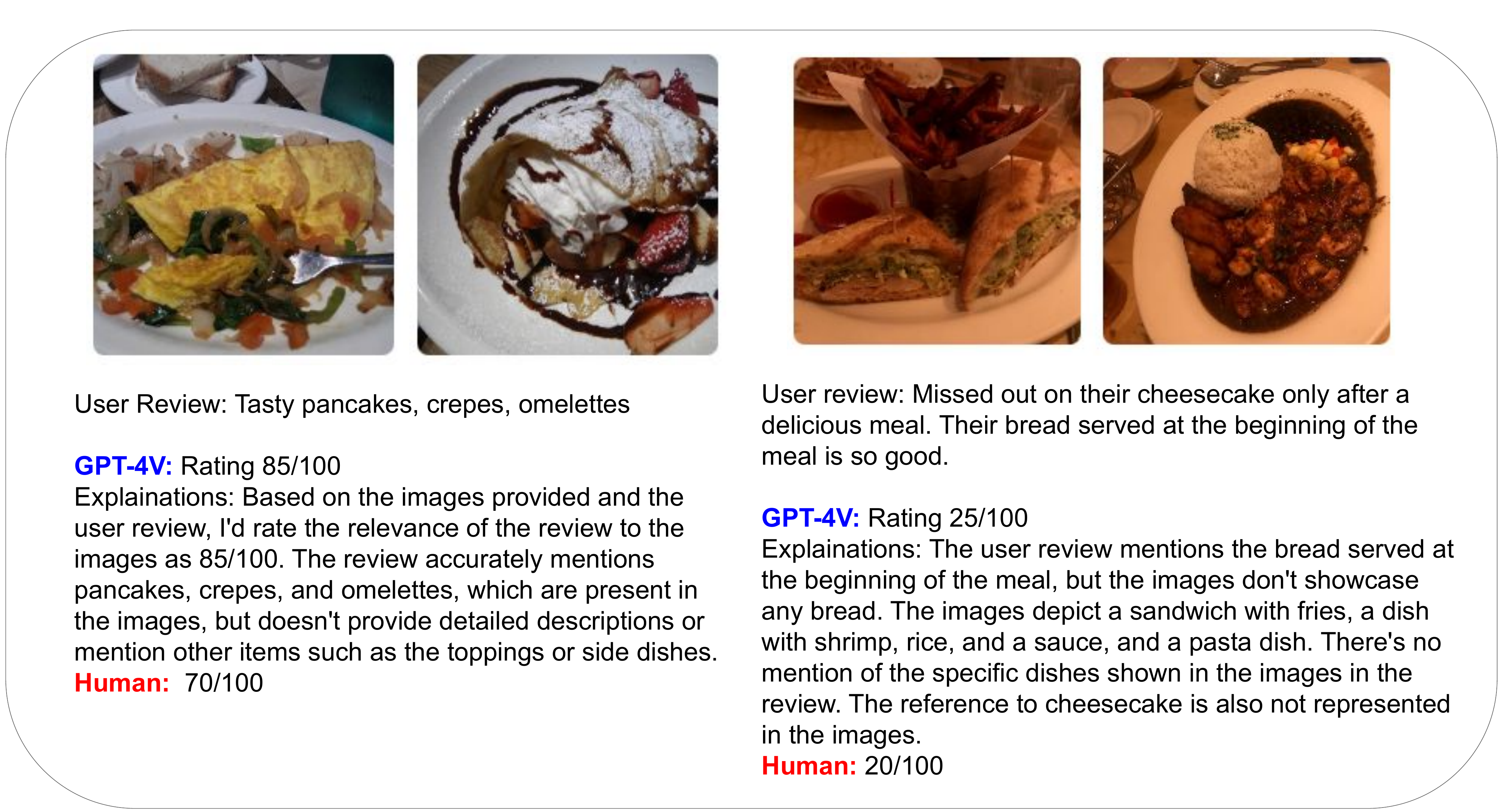} 
\caption{Examples of GPT-4V for evaluating multiple images to text task.
}
\label{fig:i2t_google_1}
\end{figure*}